# Restricted Value Iteration: Theory and Algorithms

**Weihong Zhang**                                          WZHANG@CS.WUSTL.EDU
*Department of Computer Science*
*Washington University, Saint Louis, MO 63130 USA*

**Nevin L. Zhang**                                          LZHANG@CS.UST.HK
*Department of Computer Science*
*Hong Kong University of Science & Technology*
*Clear Water Bay Road, Kowloon, Hong Kong, CHINA*

## Abstract

Value iteration is a popular algorithm for finding near optimal policies for POMDPs. It is inefficient due to the need to account for the entire belief space, which necessitates the solution of large numbers of linear programs. In this paper, we study value iteration restricted to belief subsets. We show that, together with properly chosen belief subsets, restricted value iteration yields near-optimal policies and we give a condition for determining whether a given belief subset would bring about savings in space and time. We also apply restricted value iteration to two interesting classes of POMDPs, namely informative POMDPs and near-discernible POMDPs.

## 1. Introduction

Partially Observable Markov Decision Processes (POMDPs) provide a general framework for sequential decision-making tasks where the effects of an agent's actions are nondeterministic and the states of the world or environment are not known with certainty. Due to the model generality, POMDPs have found a variety of potential applications in reality (Monahan, 1982; Cassandra, 1998b). However, solving POMDPs is computationally intractable. Extensive efforts have been devoted to developing efficient algorithms for finding solutions to POMDPs (Parr & Russell, 1995; Cassandra, Littman, & Zhang, 1997; Cassandra, 1998a; Hansen, 1998; Zhang, 2001).

Value iteration is a popular algorithm for solving POMDPs. Two central concepts in value iteration are *belief state* and *value function*. A belief state, a probability distribution over the state space, measures the probability that the environment is in each state. All possible belief states constitute a *belief space*. A value function specifies a payoff or cost for each belief state in the belief space. Value iteration proceeds in an iterative fashion. Each iteration, referred to as a *dynamic programming* (DP) update, computes a new value function from the current one. When the algorithm terminates, the final value function is used for the agent's action selection. Value iteration is computationally expensive because, at each iteration, it updates the current value function over the entire belief space, which necessitates the solution of a large number of linear programs.

One generic strategy to accelerate value iteration is to restrict value iteration, that is, DP updates, to a subset of the belief space. For simplicity, a subset of the belief space is referred to as *belief subset*. Existing value iteration algorithms working with belief subsets include a family of grid-based algorithms where DP updates calculate values for a finite grid (Lovejoy, 1991; Hauskrecht, 1997; Zhou & Hansen, 2001), and several (maybe anytime) algorithms where DP updates calculate values for a growing belief subset (Dean, Kaelbling, Kirman, & Nicholson, 1993; Washington, 1997; Hansen & Ziberstein, 1998; Hansen, 1998; Bonet & Geffner, 2000). By restricting value iteration into a belief subset, the complexity of value functions is reduced and also DP updates are more efficient. These advantages have been observed by several researchers (Hauskrecht & Fraser, 1998; Roy & Gordon, 2002; Zhang & Zhang, 2001b; Pineau, Gordon, & Thrun, 2003).

A fundamental issue in restricted value iteration is how to select a belief subset. The efficiency of value iteration and the quality of its generated value functions strongly depend on the selected belief subset. In one extreme case, if the subset is chosen to be a singleton set, value iteration is efficient, but the quality of value functions can be arbitrarily poor. In the other extreme case, if the subset is the belief space, the quality of value functions is retained, but the algorithm is inefficient. There exists a tradeoff between the size of the belief subset and the quality of value functions.

In this paper, we show that it is indeed possible for value iteration to not only work with a belief subset but also retain the quality of value functions. This is achieved by deliberately selecting a belief subset for value iteration. Sometimes, we refer to the algorithm working with our selected belief subset as *subset value iteration*. (For distinction, *restricted value iteration* refers to value iteration working with any belief subset.) The efficiency of subset value iteration depends on the size of the selected subset. We characterize a condition to *a priori* determine whether the subset is proper [1] with respect to the belief space for a given POMDP. If this is the case, subset value iteration carries the space and time advantages.

We also study two special POMDP classes, namely informative POMDPs and near-discernible POMDPs. An informative POMDP assumes that an agent has a good albeit imperfect idea about world states at any time point. For an informative POMDP, there exists a natural belief subset so that value iteration restricted to it can be more efficient than standard value iteration (Zhang & Liu, 1997). A near-discernible POMDP assumes that an agent has a good idea about world states once in a while. For a near-discernible POMDP, we propose a restricted value iteration algorithm that starts with a small belief subset and grows it gradually. The algorithm terminates as a proper tradeoff between size of the subset and policy quality is found. Because of near-discernibility, the algorithm is able to find a good tradeoff before the subset grows too large.

The algorithms developed in this paper have been tested in a variety of small maze problems designed to possess various properties as desired, and a number of problems adapted from existing research or created from our office environment. Our results show that by exploiting problem characteristics, restricted value iterations can solve larger POMDPs than standard value iteration. We show how the algorithmic performances vary with the properties of the selected belief subset for the maze problems. These small problems facilitate exposition of the properties of the chosen belief subsets. Meanwhile, the experiments provide clues on which POMDP classes are amenable to perspective algorithms.

The rest of the paper is organized as follows. In the next section, we introduce the POMDP model and value iteration. In the two subsequent sections, we present our subset value iteration algorithm and analyze its theoretical properties. In particular, in Section 3, we show how to select the belief subset and how the selected subset is related to the belief space. In Section 4, we describe the subset value iteration algorithm and discuss why it is able to achieve near optimality. In Section 5, we examine informative POMDPs and show how the algorithm that exploits the informativeness is related to the general subset value iteration (Zhang & Liu, 1997). In Section 6, we examine near-discernible POMDPs and develop an anytime algorithm. We empirically demonstrate that the algorithm is able to compute value functions of high quality. In Section 7, we survey related work to this research.

---

1. Set $A$ is a *proper* subset of set $B$ if (1) $A$ is a subset of $B$, and (2) there exists at least one element in $B$ such that it does not belong to $A$.

## 2. POMDPs and Value Iteration

This section gives a brief overview of the POMDP model and value iteration.

### 2.1 POMDPs

A POMDP is a sequential decision model for an agent who acts in a stochastic environment with only partial knowledge about the state of its environment. The set of possible states of the environment is referred to as the *state space* and is denoted by $\mathcal{S}$. At each point in time, the environment is in one of the possible states. The agent does not directly observe the state. Rather, it receives an observation about it. We denote the set of all possible observations as $\mathcal{Z}$. After receiving the observation, the agent chooses an action from a set $\mathcal{A}$ of possible actions and executes that action. Thereafter, the agent receives an immediate reward and the environment evolves stochastically into a next state.

Mathematically, a POMDP is specified by: three sets $\mathcal{S}$, $\mathcal{Z}$, and $\mathcal{A}$; a *reward function* $r(s, a)$ for $s$ in $\mathcal{S}$ and $a$ in $\mathcal{A}$; a *transition probability function* $P(s'|s, a)$; and an *observation probability function* $P(z|s', a)$ for $z$ in $\mathcal{Z}$ and $s'$ in $\mathcal{S}$. The reward function characterizes the dependency of the immediate reward on the current state $s$ and the current action $a$. The transition probability characterizes the dependency of the next state $s'$ on the current state $s$ and the current action $a$. The observation probability characterizes the dependency of the observation $z$ at the next time point on the next state $s'$ and the current action $a$.

### 2.2 Policies and Value Functions

Since the current observation does not necessarily fully reveal the identity of the current state, the agent needs to consider all previous observations and actions when choosing an action. Information about the current state contained in the current observation, previous observations, and previous actions can be summarized by a probability distribution over the state space (Aström, 1965). The probability distribution is sometimes called a *belief state* and denoted by $b$. For any possible state $s$, $b(s)$ is the probability that the current state is $s$. The set of all possible belief states is called the *belief space*. We denote it by $\mathcal{B}$.

A *policy* prescribes an action for each possible belief state. In other words, it is a mapping from $\mathcal{B}$ to $\mathcal{A}$. Associated with a policy $\pi$ is its *value function* $V^\pi$. For each belief state $b$, $V^\pi(b)$ is the expected total discounted reward that the agent receives by following the policy starting from $b$, i.e., $V^\pi(b) = E_{\pi,b}[\sum_{t=0}^{\infty} \lambda^t r_t]$ where $r_t$ is the reward received at time $t$ and $\lambda$ ($0 \le \lambda < 1$) is the *discount factor*. It is known that there exists a policy $\pi^*$ such that $V^{\pi^*}(b) \ge V^\pi(b)$ for any other policy $\pi$ and any belief state $b$ (Puterman, 1994).

Such a policy is called an *optimal policy*. The value function of an optimal policy is called the *optimal value function*. We denote it by $V^*$. For any positive number $\epsilon$, a policy $\pi$ is *$\epsilon$-optimal* if $V^\pi(b) + \epsilon \geq V^*(b)$ for any $b$ in $\mathcal{B}$.

### 2.3 Value Iteration

To explain value iteration, we need to consider how the belief state evolves over time. Let $b$ be the current belief state. The belief state at the next point in time is determined by the current belief state, the current action $a$, the next observation $z$. We denote it by $\tau(b, a, z)$. For any state $s'$, $\tau(b, a, z)$ is given by

$$\tau(b, a, z)(s') = \frac{\sum_s P(z, s'|s, a)b(s)}{P(z|b, a)}, \tag{1}$$

where $P(z, s'|s, a) = P(z|s', a)P(s'|s, a)$ and $P(z|b, a) = \sum_{s, s'} P(z, s'|s, a)b(s)$ is the renormalization constant. As the notation suggests, the constant can also be interpreted as the probability of observing $z$ after taking action $a$ in belief state $b$.

With the concept of belief state, a POMDP model can be transformed into a *belief space MDP* as follows.

- The state space is $\mathcal{B}$ and the action space is $\mathcal{A}$.

- Given a belief state $b$ and an action $a$, the transition model specifies the transition probability as follows.

$$P(b'|b, a) = \begin{cases} P(z|b, a) & \text{if } b' = \tau(b, a, z) \text{ for some } z, \\ 0 & \text{otherwise.} \end{cases}$$

- Given a belief state $b$ and action $a$, the reward model specifies immediate reward $r(b, a)$ as $r(b, a) = \sum_{s \in \mathcal{S}} b(s)r(s, a)$.

Due to this reformulation, the task of solving a POMDP can be accomplished by solving the reformulated MDP. It has been proven that the reformulated MDP has a stationary optimal policy, which can be found by stochastic dynamic programming (Bellman, 1957; Puterman, 1994).

Value iteration is a dynamic programming algorithm for finding $\epsilon$-optimal policies for an MDP. It starts with an initial value function $V_0$ and iterates using the following formula:

$$V_{n+1}(b) = \max_a [r(b, a) + \lambda \sum_z P(z|b, a)V_n(\tau(b, a, z))] \quad \forall b \in \mathcal{B} \tag{2}$$

where $V_n$ is referred to as the *$n$th-step value function*. It is known that $V_n$ geometrically converges to $V^*$ as $n$ goes to infinity.

For a given value function $V$, a policy $\pi$ is said to be *$V$-improving* if

$$\pi(b) = \arg \max_a [r(b, a) + \lambda \sum_z P(z|b, a)V(\tau(b, a, z))] \quad \forall b \in \mathcal{B}. \tag{3}$$

The following theorem tells one when to terminate value iteration given a precision requirement $\epsilon$ (Puterman, 1994). The stopping criterion depends on the quantity $\max_{b \in \mathcal{B}} |V_n(b) - V_{n-1}(b)|$, which is the maximum difference between $V_n$ and $V_{n-1}$ over the belief space. The quantity is often called *Bellman residual* between $V_n$ and $V_{n-1}$ (Puterman, 1994).

**Theorem 1** *If* $\max_b |V_n(b) - V_{n-1}(b)| \leq \epsilon(1-\lambda)/(2\lambda)$, *then the* $V_{n-1}$-*improving policy is* $\epsilon$-*optimal.*

Since there are infinitely many belief states, value functions cannot be explicitly represented. Fortunately, value functions that one encounters in the process of value iteration admit implicit finite representations (Sondik, 1971).

## 2.4 Technical and Notational Considerations

For convenience, we view functions over the state space as vectors of size $|\mathcal{S}|$. We use lower case Greek letters $\alpha$ and $\beta$ to refer to vectors and script letters $\mathcal{V}$ and $\mathcal{U}$ to refer to sets of vectors. In contrast, the upper case letters $V$ and $U$ always refer to value functions, that is functions over the belief space $\mathcal{B}$.

A set $\mathcal{V}$ of vectors *induces* a piecewise linear convex value function (say $f$) as follows: $f(b) = \max_{\alpha \in \mathcal{V}} \alpha \cdot b$ for any $b$ in $\mathcal{B}$ where $\alpha \cdot b$ is the inner product of $\alpha$ and $b$. For convenience, we shall abuse notation and use $\mathcal{V}$ to denote both a set of vectors and the value function induced by the set. Under this convention, the quantity $f(b)$ can be written as $\mathcal{V}(b)$.

A vector in a set is *useless* if its removal does not affect the function that the set induces. It is *useful* otherwise. A set of vectors is *minimal* if it contains no useless vectors. Let $\alpha$ be a vector in set $\mathcal{V}$. It is known that $\alpha$ is useful if and only if there is at least one belief state $b$ such that $\alpha \cdot b > \alpha' \cdot b$, $\forall \alpha' \in \mathcal{V} \backslash \{\alpha\}$. Such a belief state is called a *witness point* of $\alpha$ because it testifies to the fact that $\alpha$ is useful (Kaelbling, Littman, & Cassandra, 1998). To determine the usefulness of a vector in a set, it is sufficient to solve one linear program. To compute a minimal set for a given set $\mathcal{V}$ of vectors, it is sufficient to solve $|\mathcal{V}|$ linear programs. The procedure of computing a minimal set for a given set of vectors is often referred to as *pruning* a set.

## 2.5 Finite Representation of Value Functions and Value Iteration

A value function $V$ is *represented* by a set of vectors if it equals the value function induced by the set. When a value function is representable by a finite set of vectors, there is a unique minimal set that represents the function (Littman, Cassandra, & Kaelbling, 1995).

Sondik (1971) has shown that if a value function is representable by a finite set of vectors, then so are the subsequent value functions derived by DP updates. The process of obtaining the minimal representation for $V_{n+1}$ from the minimal representation of $\mathcal{V}_n$ is usually referred to as *dynamic programming (DP) update.*

In practice, value iteration for POMDPs is not carried out directly in terms of value functions themselves. Rather, it is carried out in terms of sets of vectors that represent the value functions. One begins with an initial set of vectors $\mathcal{V}_0$ (often set to a zero-vector). At each iteration, one performs a DP update on the previous minimal set $\mathcal{V}_n$ of vectors and obtains a new minimal set $\mathcal{V}_{n+1}$ of vectors. One continues until the Bellman residual $\max_b |\mathcal{V}_{n+1}(b) - \mathcal{V}_n(b)|$, which is determined by solving a sequence of linear programs, falls below a threshold.

## 3. Belief Subset Selection

In this section, we show how to select a belief subset for value iteration. We describe a condition determining whether the selected subset is proper w.r.t. the belief space. In addition, we discuss the minimal representation of value functions w.r.t. the selected subset. In the next section, we develop the subset value iteration algorithm and show why it is able to achieve near optimality.

### 3.1 Subset Selection

Our belief subset selection rests on belief updating. Let the agent's current belief be $b$. Its next belief state is $\tau(b, a, z)$ if it performs action $a$ and receives observation $z$. If we vary the belief state $b$ in the belief space $\mathcal{B}$, we obtain a set $\{\tau(b, a, z)|b \in \mathcal{B}\}$. Abusing our notation, we denote this set by $\tau(\mathcal{B}, a, z)$. In words, no matter which belief state the agent starts with, if it receives $z$ after performing $a$, its next belief state must be in $\tau(\mathcal{B}, a, z)$.

The union $\cup_{a,z} \tau(\mathcal{B}, a, z)$ takes into account the sets of belief states for all possible combinations of actions and observations. It contains all the belief states that the agent can encounter. In other words, the agent's belief state at any time point must belong to this set regardless of its initial belief state, performed actions and received observations. We denote the set by $\tau(\mathcal{B}, \mathcal{A}, \mathcal{Z})$ or simply $\tau(\mathcal{B})$. It is a *closed* set in the sense that no action can lead the agent to belief states outside $\tau(\mathcal{B})$ if the agent starts with a belief state in it. Furthermore, any belief subset between the set $\tau(\mathcal{B})$ and the belief space $\mathcal{B}$ is closed.

**Lemma 1** *The set $\tau(\mathcal{B})$ is closed. Moreover, if $\tau(\mathcal{B}) \subseteq \mathcal{B}' \subseteq \mathcal{B}$, $\mathcal{B}'$ is closed.*

As is apparent, the set $\tau(\mathcal{B})$ is a subset of the belief space $\mathcal{B}$. Its definition is an application of reachability analysis (Boutilier, Brafman, & Geib, 1998; Dean et al., 1993). Under the terminology in reachability analysis, the subset $\tau(\mathcal{B}, a, z)$ comprises the one-step reachable belief states if the agent performs action $a$ and receives observation $z$, while the subset $\tau(\mathcal{B})$ comprises the one-step reachable belief states regardless of performed actions and received observations. Although the belief subset $\tau(\mathcal{B})$ is the set of one-step reachable belief states, an appealing property, to be shown in the next subsection, is that value iteration working with it can preserve the quality of the generated value functions.

### 3.2 Subset Representation

Subset representation addresses how to represent the subsets $\tau(\mathcal{B}, a, z)$ and $\tau(\mathcal{B})$. For this, we introduce the concept of belief simplex.

**Definition 1** *Let $B = \{b_1, b_2, ..., b_k\}$ be a set of belief states. A belief simplex $\Psi$ generated by $B$ is the set of belief states $\{\sum_{i=1}^{k} \lambda_i b_i | \lambda_i \geq 0 \text{ and } \sum_{i=1}^{k} \lambda_i = 1.0\}$.*

The set $B$ is said to be a *basis* of the belief simplex $\Psi$. From the definition, the belief simplex (or simply *simplex*) is the set of convex combinations of the belief states in the basis. Following the standard terms in linear algebra, we can also talk about the *minimum basis* of a simplex. For convenience, we use notation $B_\Psi$ to denote a basis of a given simplex $\Psi$. Additionally, the simplex with the basis $\{b_1, b_2, \cdots, b_k\}$ is denoted by $\Psi(b_1, b_2, \cdots, b_k)$.

Our result is that for any $a$ and $z$, the subset $\tau(\mathcal{B}, a, z)$ is a simplex. The intuition follows. Let the number of states in a POMDP be $n$. For each $i \in \{1, 2, \ldots, n\}$, $b_i$ is a unit vector, i.e., $b_i(s)$ equals 1.0 for $s = i$ and 0.0 otherwise. For belief state $b_i$, if $P(z|b_i, a) > 0$, $\tau(b_i, a, z)$ is a belief state in $\tau(\mathcal{B}, a, z)$; if $P(z|b_i, a) = 0$, by the belief update equation $\tau(b_i, a, z)$ is undefined. In the belief space $\mathcal{B}$, it is trivial to note that any belief state can be represented as a convex combination of belief states in $\{b_1, b_2, \cdots, b_n\}$. Correspondingly, in the belief subset $\tau(\mathcal{B}, a, z)$, any belief state can be represented as a convex combination of belief states in $\{\tau(b_i, a, z) | P(z|b_i, a) > 0\}$. Hence, $\{\tau(b_i, a, z) | P(z|b_i, a) > 0\}$ is a basis of $\tau(\mathcal{B}, a, z)$. For convenience, we denote such a basis by $B_{\tau(\mathcal{B}, a, z)}$.

**Theorem 2** *For any pair $[a, z]$, the subset $\tau(\mathcal{B}, a, z)$ is a simplex.*

**Proof:** See Appendix A. $\qquad\qquad\square$

By the above theorem, the subset $\tau(\mathcal{B})$ is a union of simplices. Although the subset $\tau(\mathcal{B})$ is not linearly representable in its own, it is the union of linearly representable sets. Later in the section, this property is crucial to and will be exploited in finding the minimal representing sets of value functions w.r.t. the belief subset $\tau(\mathcal{B})$.

To concretize the ideas on subset representation, we give a POMDP example and visualize the simplices for actions and observations. Before presenting the example, we mention that we shall use it for additional purposes later on in this paper. First, we shall use it to show the difference between two conditions determining whether the subset $\tau(\mathcal{B})$ is a proper subset of the belief space. Second, we shall use it to demonstrate the fundamental differences between two restricted value iteration algorithms.

**Example** The POMDP has three states $\{s_1, s_2, s_3\}$, two actions $\{a_1, a_2\}$ and two observations $\{z_1, z_2\}$. We define the transition and observation model for action $a_1$. These models for $a_2$ can be defined similarly. To shorten notations, we use $p_{ij}$ to denote the transition probability $P(s_j|s_i, a_1)$ and $q_{ij}$ to denote the observation probability $P(z_j|s_i, a_1)$. We assume that (1) for any state $s_i$, the probability $p_{i1}$ is equal to $p_{i2}$, i.e., $p_{i1} = p_{i2}$; (2) at each state $s_i$, observations $z_1$ and $z_2$ are received with the same probability, i.e., $q_{i1} = q_{i2} = 0.5$ for each $i$; and (3) $p_{11} > p_{21} > p_{31}$. Under these assumptions, the matrix

$$P_{a_1 z_1} = \begin{pmatrix} 0.5 p_{11} & 0.5 p_{21} & 0.5 p_{31} \\ 0.5 p_{11} & 0.5 p_{21} & 0.5 p_{31} \\ 0.5(1 - 2p_{11}) & 0.5(1 - 2p_{21}) & 0.5(1 - 2p_{31}) \end{pmatrix}. \qquad (4)$$

Because of the first assumption, the first two rows of the matrix are the same. In the third row, the probability $p_{i3}$ is replaced with $1 - p_{i1} - p_{i2}$, i.e., $1 - 2p_{i1}$.

We compute the basis of the belief subset $\tau(\mathcal{B}, a_1, z_1)$. Let the basis of the belief space $\mathcal{B}$ be the set $\{(1.0, 0, 0)^T, (0, 1.0, 0)^T, (0, 0, 1.0)^T\}$. (For a matrix or vector $A$, $A^T$ denotes its transpose.) For action $a_1$ and observation $z_1$, the next belief states, denoted by $A_i$s, are:

$$\begin{aligned} A_1 &= \tau((1.0, 0, 0)^T, a_1, z_1) = (p_{11}, p_{11}, 1.0 - 2p_{11})^T \\ A_2 &= \tau((0, 1.0, 0)^T, a_1, z_1) = (p_{21}, p_{21}, 1.0 - 2p_{21})^T \\ A_3 &= \tau((0, 0, 1.0)^T, a_1, z_1) = (p_{31}, p_{31}, 1.0 - 2p_{31})^T \end{aligned} \qquad (5)$$

Interestingly, it can be shown that $A_2$ is a convex combination of $A_1$ and $A_3$. In fact, in can be verified that

$$\tau((0, 1.0, 0)^T, a_1, z_1) = \lambda_1 \tau((1.0, 0, 0)^T, a_1, z_1) + \lambda_2 \tau((0, 0, 1.0)^T, a_1, z_1)$$

where $\lambda_1 = \frac{p_{21} - p_{31}}{p_{11} - p_{31}}$ and $\lambda_2 = \frac{p_{11} - p_{21}}{p_{11} - p_{31}}$. Thus $\lambda_1 + \lambda_2 = 1.0$. Our third assumption ensures that $\lambda_1$ and $\lambda_2$ are greater than 0.0. Because $A_2$ is a convex combination of $A_1$ and $A_3$, the three belief states $A_1$ to $A_3$ lie in the same straight line.

Figure 1 visualizes the belief space $\mathcal{B}$ in the left and the simplex $\tau(\mathcal{B}, a_1, z_1)$ in the right. The right chart is based on these parameters: $p_{11} = 0.5$, $p_{21} = 0.4$ and $p_{31} = 0.1$. The belief states $A_i$s are the following: $A_1 = (0.5, 0.5, 0.0)^T$, $A_2 = (0.4, 0.4, 0.2)^T$ and $A_3 = (0.1, 0.1, 0.8)^T$. We see that the belief space is a triangle area and the belief simplex is a line segment in that area. In the next subsection, we shall return to this point and show why.

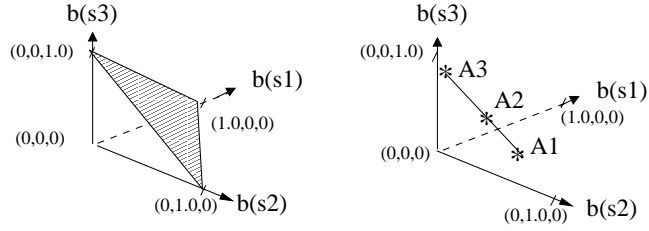

Figure 1: A graphical representation of belief space and belief simplex. See text for explanations.

To show the belief subset $\tau(\mathcal{B})$, we continue to define the transition and the observation models for action $a_2$.[2] We may follow the three assumptions for $a_1$ to define these models for $a_2$, but give a different set of transition probabilities so that $a_2$ differs from $a_1$.

By the second assumption that observation $z_1$ and $z_2$ are received with the same probability, the matrix $P_{a_i, z_1}$ is identical to $P_{a_i, z_2}$ for $i \in \{1, 2\}$. So, the simplex $\tau(\mathcal{B}, a_i, z_1)$ is identical to $\tau(\mathcal{B}, a_i, z_2)$ for each $i$. As such, the subset $\tau(\mathcal{B})$ consists of two line segments in the entire belief space. □

## 3.3 Belief Subset and Belief Space

We discuss the relationship between the set $\tau(\mathcal{B})$ and the belief space $\mathcal{B}$. Since the set $\tau(\mathcal{B})$ is a union of simplices, it helps to show how each simplex $\tau(\mathcal{B}, a, z)$ is related to the belief space $\mathcal{B}$. For an action $a$ and observation $z$, it turns out that a matrix derived from the transition and observation models plays a central role in determining the simplex. Such a matrix, denoted by $P_{az}$, is of dimension $|\mathcal{S}| \times |\mathcal{S}|$ and its entry at $(s, s')$ is the joint probability $P(s', z|s, a)$, i.e.,

$$P_{az} = \begin{pmatrix} P(s'_1, z|s_1, a) & P(s'_1, z|s_2, a) & \cdots & P(s'_1, z|s_n, a) \\ P(s'_2, z|s_1, a) & P(s'_2, z|s_2, a) & \cdots & P(s'_2, z|s_n, a) \\ \cdots & \cdots & \cdots & \cdots \\ P(s'_n, z|s_1, a) & P(s'_n, z|s_2, a) & \cdots & P(s'_n, z|s_n, a) \end{pmatrix}.$$

---

2. Other components of the POMDP do not affect our discussions here and are omitted for convenience.

The matrix can be used to relate the next belief $\tau(b, a, z)$ and the current $b$. If $b$ and $\tau(b, a, z)$ are viewed in column vector form, the belief update Equation (1) can be rewritten as $\tau(b, a, z) = \frac{1}{p(z|b,a)} P_{az} b$. Hence, $\tau(b, a, z)$ is a transformation of $b$ and the matrix $P_{az}$ is called the *transformational matrix*. The following lemma characterizes a condition under which the simplex $\tau(\mathcal{B}, a, z)$ is the same as the belief space $\mathcal{B}$.

**Lemma 2** *For any $[a, z]$, there exists a bijection between the simplex $\tau(\mathcal{B}, a, z)$ and the space $\mathcal{B}$ if the matrix $P_{az}$ is invertible* [3].

This can be seen from the fact $\tau(b, a, z) = \frac{1}{p(z|b,a)} P_{az} b$ and $\tau(\mathcal{B}, a, z)$ is the set of the transformed belief states from the belief space. Consequently, the simplex $\tau(\mathcal{B}, a, z)$ is a *proper* subset of $\mathcal{B}$ if the matrix $P_{az}$ is degenerate. We note that the matrix $P_{az}$ is degenerate if there exists a state $s'$ such that $P(z|s', a) = 0.0$, i.e., there is a state such that the agent can never receive the observation $z$ (if action $a$ is performed). This is because all the entries in the row corresponding to $s'$ are $0.0$ in the matrix. In this case, by the lemma, the set $\tau(\mathcal{B}, a, z)$ must be a proper subset of the belief space $\mathcal{B}$.

**Corollary 1** *If there is a state $s$ such that $P(z|s, a) = 0.0$, the set $\tau(\mathcal{B}, a, z)$ is a proper subset of the belief space $\mathcal{B}$.*

However, that $\tau(\mathcal{B}, a, z)$ is a proper subset does not necessarily imply that there exists a state $s$ such that $P(z|s, a) = 0.0$. Consequently, to determine if a belief simplex is proper, the above corollary provides a sufficient condition; in contrast, Lemma 2 provides a sufficient and necessary condition. This will be illustrated by continuing our discussions on the POMDP example.

**Example (Continued)** We consider the simplex $\tau(\mathcal{B}, a_1, z_1)$. By the second assumption, $z_1$ can be observed at any state. So, there does not exist a state $s$ such that $P(z_1|s, a_1) = 0.0$. On the other hand, the matrix $P_{a_1 z_1}$ is degenerate because it has the same two rows. By Lemma 2, the simplex $\tau(\mathcal{B}, a_1, z_1)$ is a proper set of the belief space. In fact, as seen in Figure 1, it is a line segment, which can be viewed as a degenerate belief space. $\qquad\square$

We proceed to discover the relationship between the belief set $\tau(\mathcal{B}, a, z)$ and the belief space $\mathcal{B}$. Since $\tau(\mathcal{B}, a, z)$ is the union of the simplices $\tau(\mathcal{B}, a, z)$, it is a proper subset of the belief space if so is each simplex. In turn, this requires that each transformational matrix is degenerate.

**Theorem 3** *The subset $\tau(\mathcal{B})$ is a proper subset of belief space $\mathcal{B}$ only if the transformational matrices for all actions and observations are degenerate.*

### 3.4 Subset Value Functions

We discuss value functions whose domains are belief subset $\tau(\mathcal{B}, a, z)$ or $\tau(\mathcal{B})$. For simplicity, we refer to them as *subset value functions*. The problem we will examine is, given a set of vectors representing a subset value function, how to compute a minimal set w.r.t. a belief subset. We first consider the case where the subset is a simplex.

---

3. A matrix is invertible if its determinant is non-zero. It is degenerate otherwise.

In order to calculate a minimal set of vectors, one needs to determine the usefulness of a vector in a set w.r.t. the simplex. Let $\beta$ be a vector in set $\mathcal{V}$ and the simplex be $\tau(\mathcal{B}, a, z)$. The vector $\beta$ is useful w.r.t. $\tau(\mathcal{B}, a, z)$ if and only if there is a belief state $b$ in the simplex such that $\beta \cdot b \geq \alpha \cdot b + x$ where $x$ is a sufficiently small positive number and $\alpha$ is a vector in the set $\mathcal{V} - \{\beta\}$. Moreover, if such a belief state $b$ exists, since it is in the simplex, $b$ must be representable by the belief states in the basis $B_{\tau(\mathcal{B}, a, z)}$, i.e., $b = \sum_i \lambda_i \tau(b_i, a, z)$. If we replace $b$ by $\sum_i \lambda_i \tau(b_i, a, z)$ in $\beta \cdot b \geq \alpha \cdot b + x$, the condition of determining $\beta$'s usefulness is equivalent to this: whether there exists a series of nonnegative numbers $\lambda_i$s such that for any vector $\alpha$ in $\mathcal{V}$,

$$\beta \cdot \sum_i \lambda_i \tau(b_i, a, z) \geq \alpha \cdot \sum_i \lambda_i \tau(b_i, a, z) + x.$$

Rewriting the above inequality, we have

$$\sum_i \left[ \beta \cdot \tau(b_i, a, z) \right] \lambda_i \geq \sum_i \left[ \alpha \cdot \tau(b_i, a, z) \right] \lambda_i + x. \qquad (6)$$

To determine $\beta$'s usefulness, the procedure `simplexLP` in Table 1 is used. When the optimality of the linear program is reached, one checks its objective $x$. If it is positive, there exists a belief state in belief simplex $\tau(\mathcal{B}, a, z)$ such that at this belief state $\beta$ dominates other vectors. The belief state is a witness point of $\beta$. It is represented as $\sum_i \lambda_i \tau(b_i, a, z)$ where $\lambda_i$s are the solutions (values of variables) of the linear program. In this case, the belief state $\sum_i \lambda_i \tau(b_i, a, z)$ is returned. For other cases, no belief state is returned and $\beta$ is a useless vector.

---

`simplexLP`$(\beta, \mathcal{V}, B_{\tau(\mathcal{B}, a, z)})$:
1. Variables: $x$, $\lambda_i$ for each $i$
2. Maximize: $x$.
3. Constraints:
4.     $\sum_i \left[ \beta \cdot \tau(b_i, a, z) \right] \lambda_i \geq \sum_i \left[ \alpha \cdot \tau(b_i, a, z) \right] \lambda_i + x$ for $\forall \alpha \in \mathcal{V} - \{\beta\}$
           and for each $i$, $\tau(b_i, a, z) \in B_{\tau(\mathcal{B}, a, z)}$
5.     $\sum_i \lambda_i = 1$, $\lambda_i \geq 0$ for $i$.

`simplexPrune`$(\mathcal{V}, B_{\tau(\mathcal{B}, a, z)})$:
1. $\mathcal{U} \leftarrow \mathcal{V}$, $\mathcal{V} \leftarrow \emptyset$
2. **For** each $\beta$ in $\mathcal{U}$
3.     $b \leftarrow$ `simplexLP`$(\beta, \mathcal{U}, B_{\tau(\mathcal{B}, a, z)})$
4.     **If** $b \neq$ `null`
5.         $\mathcal{V} \leftarrow \mathcal{V} \cup \{\beta\}$
6. **Return** $\mathcal{V}$

---

Table 1: The procedure to compute the minimal set of vectors over a simplex

To determine a vector's usefulness in a given set, one linear program needs to be solved. If a vector is useless, its removal does not change the value function that the set induces.

Therefore, to compute a minimal set for a given set $\mathcal{V}$ w.r.t. a simplex, one needs to solve $|\mathcal{V}|$ linear programs.

This procedure is implemented in `simplexPrune` of Table 1. Its input has two arguments: a set of vectors $\mathcal{V}$ [4] and a basis of the simplex $B_{\tau(\mathcal{B},a,z)}$. A set $\mathcal{U}$ is initialized to be the set $\mathcal{V}$ and the set $\mathcal{V}$ to be empty at line 1. Useful vectors are added to the set $\mathcal{V}$ in the sequel. For each vector in set $\mathcal{U}$, at line 3 the procedure `simplexLP` is called to determine its uselessness. If it returns a belief state, the vector is added to the set $\mathcal{V}$ at line 5. Eventually, the set $\mathcal{V}$ becomes the minimal representation of $\mathcal{U}$ w.r.t. simplex $\tau(\mathcal{B}, a, z)$.

To compute a minimal set of vectors w.r.t. the subset $\tau(\mathcal{B})$, one needs to determine a vector's usefulness w.r.t. the subset. In turn, one needs to determine its usefulness w.r.t. each simplex. Again, let the set be $\mathcal{V}$ and the vector be $\beta$. If $\beta$ is useful w.r.t. a simplex, it must be useful w.r.t. the subset. However, if it is useless w.r.t. a simplex, it may be useful w.r.t. another simplex. Hence, for a vector, if $\beta$ has been identified as useful, there is no need to check it again for subsequent simplices. After all the simplices have been examined, if $\beta$ is useless w.r.t. all simplices, it is useless w.r.t. the subset. By removing all useless vectors w.r.t. the subset, one obtains the minimal set.

## 4. Subset Value Iteration

In this section, we first describe the value iteration algorithm in belief subset $\tau(\mathcal{B})$. We then show that the algorithm is able to achieve near optimality. Finally, we analyze its complexity and report empirical studies.

### 4.1 Belief Subset MDP

Because the subset $\tau(\mathcal{B})$ is closed, we are able to define a so-called *belief subset MDP* (or simply *subset MDP*). Its state space is the chosen subset $\tau(\mathcal{B})$ and other components are the same as those in the MDP transformed from the original POMDP (Section 2.3). The only difference between the two MDPs lies in their state spaces: the state space of the belief subset MDP is a subset of the state space of the belief space MDP.

### 4.2 Subset DP Updates

By MDP theory, the subset MDP admits the following DP update equation where $V_n^{\tau(\mathcal{B})}$ represents its $n$th-step value function.

$$V_{n+1}^{\tau(\mathcal{B})}(b) = \max_a \{ r(b,a) + \lambda \sum_z P(z|b,a) V_n^{\tau(\mathcal{B})}(\tau(b,a,z)) \} \quad \forall b \in \tau(\mathcal{B}). \qquad (7)$$

Following the equation, an implicit DP update computes the minimal set $\mathcal{V}_{n+1}^{\tau(\mathcal{B})}$ representing value function $V_{n+1}^{\tau(\mathcal{B})}$ from $\mathcal{V}_n^{\tau(\mathcal{B})}$ representing $V_n^{\tau(\mathcal{B})}$. Note that the domains of value functions are belief subset $\tau(\mathcal{B})$. For simplicity, such a step is called *subset DP update*.

Implicit subset DP updates can be carried out as standard DP updates. Here, we present a two-pass algorithm due to its conceptual simplicity (Monahan, 1982). It constructs the next set of vectors in two steps – an enumeration step enumerating all possible vectors and a reduction step removing useless vectors. In the following, we focus on the enumeration step. For the reduction step, since the usefulness of a vector is w.r.t. the subset $\tau(\mathcal{B})$, the techniques in the preceding section can be used.

---

4. For simplicity, we assume that the set $\mathcal{V}$ does not contain duplicate vectors. Duplicates can be removed by a simple componentwise check.

Given a set $\mathcal{V}_n^{\tau(\mathcal{B})}$, a vector in the representing set of $V_{n+1}^{\tau(\mathcal{B})}$ can be defined by a pair of action and a mapping from the set $\mathcal{Z}$ of observations to the set $\mathcal{V}_n^{\tau(\mathcal{B})}$. To be precise, for an action $a$ and a mapping $\delta$, a vector, denoted by $\beta_{a,\delta}$, is defined as follows [5]. For each $s \in \mathcal{S}$,

$$\beta_{a,\delta}(s) = r(s,a) + \lambda \sum_z \sum_{s'} P(s'|s,a)P(z|s',a)\delta_z(s') \tag{8}$$

where $\delta_z$ is the mapped vector for observation $z$.

If we enumerate all possible combinations of actions and mappings above, we can define various vectors. These vectors form a set

$$\{\beta_{a,\delta} | a \in \mathcal{A}, \delta : \mathcal{Z} \to \mathcal{V}_n^{\tau(\mathcal{B})} \ \& \ \forall z, \delta_z \in \mathcal{V}_n^{\tau(\mathcal{B})}\}. \tag{9}$$

The set is denoted by $\mathcal{V}_{n+1}^{\tau(\mathcal{B})}$. By MDP theory, it represents the value function $V_{n+1}^{\tau(\mathcal{B})}$ if the set $\mathcal{V}_n^{\tau(\mathcal{B})}$ represents $V_n^{\tau(\mathcal{B})}$.

**Lemma 3** *The set $\mathcal{V}_{n+1}^{\tau(\mathcal{B})}$ represents value function $V_{n+1}^{\tau(\mathcal{B})}$ if $\mathcal{V}_n^{\tau(\mathcal{B})}$ represents $V_n^{\tau(\mathcal{B})}$.*

The above DP update works in a *collective* fashion in that it directly computes value functions over $\tau(\mathcal{B})$. An alternative way to conduct DP updates is to compute value functions for individual simplices one by one. The rationale is that, by letting a DP update work with the finer-grained belief subsets, it could be more efficient than its collective version. A DP update in an individual fashion constructs a collection $\{\mathcal{V}_{n+1}^{\tau(\mathcal{B},a,z)}\}$ of vector sets from a given collection $\{\mathcal{V}_n^{\tau(\mathcal{B},a,z)} | a \in \mathcal{A}, z \in \mathcal{Z}\}$ where each $\mathcal{V}_n^{\tau(\mathcal{B},a,z)}$ represents $V_n^{\tau(\mathcal{B})}$ in the simplex $\tau(\mathcal{B},a,z)$. We consider how to construct a set $\mathcal{V}_{n+1}^{\tau(\mathcal{B},a',z')}$ for one simplex $\tau(\mathcal{B},a',z')$.

Likewise, a vector $\beta_{a,\delta}$ in $\mathcal{V}_{n+1}^{\tau(\mathcal{B},a',z')}$ can be defined by an action $a$ and a mapping $\delta$. The fact that $\tau(b,a,z)$ must be in $\tau(\mathcal{B},a,z)$ for any $b$ implies that for any $z$, $\delta_z$ can be restricted to a vector in the set $\mathcal{V}_n^{\tau(\mathcal{B},a,z)}$. By altering actions and mappings, one obtains the following set:

$$\{\beta_{a,\delta} | a \in \mathcal{A}, \quad \delta : \mathcal{Z} \to \cup_{a,z} \mathcal{V}_n^{\tau(\mathcal{B},a,z)}, \ \& \ \forall z, \delta_z \in \mathcal{V}_n^{\tau(\mathcal{B},a,z)}\}. \tag{10}$$

It differs from (8) in that for an observation $z$, the mapped vector by $\delta$ is restricted to the set $\mathcal{V}_n^{\tau(\mathcal{B},a,z)}$. The above set is denoted by $\mathcal{V}_{n+1}^{\tau(\mathcal{B},a',z')}$. To obtain its minimal representation, one removes useless vectors from the set w.r.t. the simplex $\tau(\mathcal{B},a',z')$. The value function that the set $\mathcal{V}_{n+1}^{\tau(\mathcal{B},a',z')}$ induces is equal to the value function $V_n^{\tau(\mathcal{B})}$ in the belief simplex $\tau(\mathcal{B},a',z')$.

---

5. The procedure of defining a vector actually constructs an $(n+1)$th-step policy tree. (See, e.g., Zhang and Liu 1997, for details.)

**Lemma 4** *For any action $a$ and observation $z$, each set $\mathcal{V}_{n+1}^{\tau(\mathcal{B},a,z)}$ represents the same value function $V_{n+1}^{\tau(\mathcal{B})}$ in the simplex $\tau(\mathcal{B},a,z)$ if each $\mathcal{V}_n^{\tau(\mathcal{B},a,z)}$ represents $V_n^{\tau(\mathcal{B})}$ in the same simplex.*

Although subset DP updates can be carried out either collectively or individually, they are essentially equivalent in terms of value functions induced.

**Theorem 4** *Let $\mathcal{U} = \cup_{a,z} \mathcal{V}_{n+1}^{\tau(\mathcal{B},a,z)}$. For any $b \in \tau(\mathcal{B})$, $\mathcal{U}(b) = \mathcal{V}_{n+1}^{\tau(\mathcal{B})}(b)$.*

It is worthwhile to note that for two pairs of actions and observations, the simplices $\tau(\mathcal{B}, a_1, z_1)$ and $\tau(\mathcal{B}, a_2, z_2)$ might not be disjoint. A few remarks are in order for this case. First, by Theorem 4, for any $b$ in the intersection of the simplices, $\mathcal{V}_{n+1}^{\tau(\mathcal{B},a_1,z_1)}(b) = \mathcal{V}_{n+1}^{\tau(\mathcal{B},a_2,z_2)}(b)$. This is because both sets represent $V_{n+1}^{\tau(\mathcal{B})}$ in $\tau(\mathcal{B})$. Second, if a subset DP update is carried out individually, it may generate more vectors than its collective version. This is because the two sets $\mathcal{V}_{n+1}^{\tau(\mathcal{B},a_1,z_1)}$ and $\mathcal{V}_{n+1}^{\tau(\mathcal{B},a_2,z_2)}$ may contain duplicate vectors.

Finally, we note that to achieve computational savings, sophisticated algorithms for standard DP updates can be applied to subset DP updates. Let us take incremental pruning, one of the most efficient algorithms, as an example (Cassandra et al., 1997; Zhang & Liu, 1997). In standard incremental pruning, all the pruning operations are w.r.t. the belief space; however, when it is used in subset DP updates, all the pruning operations are w.r.t. the belief subsets.

## 4.3 Analysis

We analyze several theoretical properties of the subset value iteration algorithm. Our main results include: value functions generated by subset value iteration are equivalent to those by standard value iteration in some sense; to achieve near optimality, value iteration needs to account for at least the belief subset $\tau(\mathcal{B})$; the value function generated by subset value iteration can be used for near optimal decision-making in the entire belief space if the algorithm is appropriately terminated.

### 4.3.1 Belief Subset, Value Functions and Value Iterations

Subset value iteration generates a series $\{V_n^{\tau(\mathcal{B})}\}$ of value functions. If its initial value function $V_0^{\tau(\mathcal{B})}$ is the same as the initial $V_0$ of standard value iteration in $\tau(\mathcal{B})$, subset value iteration generates the same series of value functions as standard value iteration in $\tau(\mathcal{B})$.

**Theorem 5** *If $V_0^{\tau(\mathcal{B})}(b) = V_0(b)$ for any $b$ in $\tau(\mathcal{B})$, then $V_n^{\tau(\mathcal{B})}(b) = V_n(b)$ for any $n$ and any $b$ in $\tau(\mathcal{B})$.*

**Proof:** We first consider one DP update computing value function $V_{n+1}$ from the current $V_n$ by DP Equation (2). In its right hand side, since $\tau(b, a, z)$ must belong to the subset $\tau(\mathcal{B})$, the notation $V_n(\tau(.,.,.))$ can be interpreted as a value function over subset $\tau(\mathcal{B})$ rather than belief space $\mathcal{B}$. Comparing DP Equation (2) for the belief space $\mathcal{B}$ and Equation (7) for belief subset $\tau(\mathcal{B})$, we see that $V_{n+1}$ and $V_{n+1}^{\tau(\mathcal{B})}$ represent the same value function in $\tau(\mathcal{B})$ if so do $V_n$ and $V_n^{\tau(\mathcal{B})}$.

The theorem is true for $n = 0$ by the given condition. It is true for $n > 0$ by induction. □

More interestingly, the value function $V_n^{\tau(\mathcal{B})}$ at step $n$ can be used to derive the value function $V_{n+1}$ in standard value iteration. To see why, we first note that a subset value function $V^{\tau(\mathcal{B})}$ can be used to define a value function (say $V$) by one-step lookahead operation as follows:

$$V(b) = \max_a \{ r(b, a) + \lambda \sum_z P(z|b, a) V^{\tau(\mathcal{B})}(\tau(b, a, z)) \} \quad \forall \ b \in \mathcal{B}. \tag{11}$$

The so-defined $V$ is called $V^{\tau(\mathcal{B})}$-*improving value function*. Second, comparing Equations (11) and (2), we see that the $V_n^{\tau(\mathcal{B})}$-improving value function is actually $V_{n+1}$ if $V_n^{\tau(\mathcal{B})}$ is equal to $V_n$ in $\tau(\mathcal{B})$.

Consequently, although subset value iteration works with $\tau(\mathcal{B})$, value functions generated in standard value iteration can be derived. In this sense, we say $\tau(\mathcal{B})$ is a *sufficient* belief subset since it enables subset value iteration to preserve standard value functions without "loss".

Since subset value iteration retains the quality of value functions, it can be regarded as an *exact* algorithm. One interesting question is, if value iteration intends to retain quality, can it work with a proper subset of $\tau(\mathcal{B})$? In general, the answer is no. The reason follows. To compute $V_{n+1}$, one needs to keep values $V_n^{\tau(\mathcal{B})}$ for belief states in $\tau(\mathcal{B})$. Otherwise, if one accounts for a proper subset $\mathcal{B}'$ of $\tau(\mathcal{B})$, it can be proven that there exists a belief state $b$ in $\mathcal{B}$, an action $a$ and an observation $z$ such that $\tau(b, a, z)$ does not belong to $\mathcal{B}'$. It's known that the value update of $V_{n+1}(b)$ depends on the values for all possible next belief states. Due to the unavailability of $V_n^{\tau(\mathcal{B})}(\tau(b, a, z))$, the value $V_{n+1}(b)$ cannot be calculated exactly. Consequently, if value iteration works with a proper subset of $\tau(\mathcal{B})$, it cannot be exact. In other words, it should be an *approximate* algorithm. To make it be exact, value iteration needs consider at least $\tau(\mathcal{B})$. In this sense, the subset $\tau(\mathcal{B})$ is said to be a *minimal* sufficient set.

Informally we use Figure 2 to illustrate the relationship between belief subsets and value iteration. In the figure, circles represent belief sets. The minimum belief subset for value iteration to retain quality is $\tau(\mathcal{B})$, while the maximum subset is the belief space $\mathcal{B}$ itself. If value iteration works with belief subset $\mathcal{B}'$ (denoted by dashed circles) between $\tau(\mathcal{B})$ and $\mathcal{B}$, its quality is also retained. However, if it works with a proper belief subset of $\tau(\mathcal{B})$, in general it is unable to retain the quality of value functions.

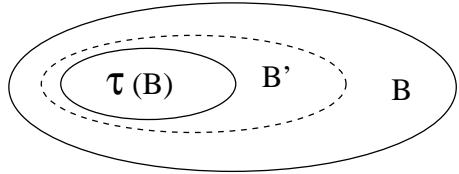

Figure 2: The relationship between belief subsets and value iteration

### 4.3.2 Stopping criterion and decision making

Subset value iteration starts with an initial value function. As it continues, the Bellman residual $\max_{b \in \tau(\mathcal{B})} |V_n^{\tau(\mathcal{B})}(b) - V_{n-1}^{\tau(\mathcal{B})}(b)|$, the maximum difference between two value functions over the subset $\tau(\mathcal{B})$, must become smaller. By MDP theory, if the quantity falls below $\epsilon(1-\lambda)/(2\lambda)$, the value function $\mathcal{V}_{n-1}^{\tau(\mathcal{B})}$ is $\epsilon$-optimal w.r.t. the subset MDP. In the following, we show that the $\epsilon$-optimality can be extended to the entire belief space by appropriately terminating the subset value iteration algorithm.

Let the output value function be $V_{n-1}^{\tau(\mathcal{B})}$. It can be used to define a policy for any belief state in $\mathcal{B}$ as in Equation (3) where $V$ is replaced by $V_{n-1}^{\tau(\mathcal{B})}$. The policy $\pi$ is said to be $V_{n-1}^{\tau(\mathcal{B})}$-improving. Note that the policy prescribes an action for any belief state in the belief space. The following theorem tells one when to terminate subset value iteration such that the $V_{n-1}^{\tau(\mathcal{B})}$-improving policy is $\epsilon$-optimal over the belief space.

**Theorem 6** *If* $\max_{b \in \tau(\mathcal{B})} |V_n^{\tau(\mathcal{B})}(b) - V_{n-1}^{\tau(\mathcal{B})}(b)| \leq \epsilon(1-\lambda)/(2\lambda^2 |\mathcal{Z}|)$, *then the* $V_{n-1}^{\tau(\mathcal{B})}$-improving *policy is* $\epsilon$-optimal over the entire belief space $\mathcal{B}$.

**Proof:** See Appendix A. □

This theorem is important for two reasons. First, although subset value iteration outputs a subset value function, $\epsilon$-optimal value functions over the entire space can be induced by one-step lookahead operation. Second, it implies that the $V_{n-1}^{\tau(\mathcal{B})}$-improving policy is $\epsilon$-optimal if the condition is met. We know that $\tau(\mathcal{B})$ consists of all possible belief states the agent encounters after the initial belief state. However, we have no assumption about the initial belief state. It may or may not belong to this set. The theorem means that the agent is still able to select a near optimal action for an initial belief state even if it is not in the subset. In fact, the agent can always select near optimal action for any belief state in the entire belief space.

Finally, we note that to guarantee the $\epsilon$-optimality, when compared with the condition in Theorem 1, subset value iteration uses a more restrictive condition. For convenience, it is sometimes called a *strict stopping criterion*. In contrast, the condition in standard value iteration is called a *loose stopping criterion*.

## 4.4 Complexity

To put in use for a POMDP, the subset value iteration algorithm would take two steps: determining if the algorithm can bring about savings in time and then running the subset value iteration if it can. The first step needs to compute $|\mathcal{A}||\mathcal{Z}|$ determinants $|P_{az}|$. Since the complexity of computing $|P_{az}|$ is $|\mathcal{S}|^3$, the first step has the complexity of $O(|\mathcal{A}||\mathcal{Z}||\mathcal{S}|^3)$. This is the polynomial part of the complexity of subset value iteration. The second step is much harder than the first step. It is known that finding the optimal policy for even a simplified finite horizon POMDP is PSPACE-complete (Papadimitriou & Tsitsiklis, 1987; Burago, de Rougemont, & Slissekno, 1996; Littman, Goldsmith, & Mundhenk, 1998). Recently, it has been proven that finding the optimal policy for an infinite-horizon POMDP is incomputable (Madani., Hanks, & Condon, 1999).

We compare the subset value iteration with standard value iteration. Standard DP updates improve values for the space $\mathcal{B}$, while subset DP updates improve values for the subset $\tau(\mathcal{B})$. If the initial set $\mathcal{V}_0^{\tau(\mathcal{B})}$ is equal to the initial set $\mathcal{V}_0$ in standard value iteration, because $\tau(\mathcal{B})$ is a subset of $\mathcal{B}$, the vectors in $\mathcal{V}_1^{\tau(\mathcal{B})}$ must be in $\mathcal{V}_1$, and $\mathcal{V}_1^{\tau(\mathcal{B})}$ is a subset of $\mathcal{V}_1$. Inductively, $\mathcal{V}_n^{\tau(\mathcal{B})}$ is a subset of $\mathcal{V}_n$ for any $n$. This analysis suggests two advantages of subset value iteration if the subset $\tau(\mathcal{B})$ is a proper subset of the belief space $\mathcal{B}$. First, fewer vectors are needed to represent a value function over a belief subset. This is the representational advantage in space. For $\mathcal{V}_{n+1}$, its size can be as large as $|\mathcal{A}||\mathcal{V}_n|^{|\mathcal{Z}|}$. For $\mathcal{V}_{n+1}^{\tau(\mathcal{B})}$, its size can be as large as $|\mathcal{A}||\mathcal{V}_n^{\tau(\mathcal{B})}|^{|\mathcal{Z}|}$. Clearly, subset DP update generates fewer vectors. Second, fewer vectors means lesser degree of time complexity since computing vectors needs to solve linear programs. This is the computational advantage in time.

However, the advantages strongly depend upon the size of the subset $\tau(\mathcal{B})$. If each simplex $\tau(\mathcal{B}, a, z)$ is the same as the belief space $\mathcal{B}$ and DP updates are conducted in an individual fashion, subset DP updates could be $|\mathcal{A}||\mathcal{Z}|$ times slower than standard DP updates. This is the worst case complexity. Fortunately, by discussions in the previous section (Theorem 3), we know that given a POMDP we are able to determine whether the selected subset $\tau(\mathcal{B})$ is a proper subset of the belief space before solving it.

Although Theorem 3 gives a condition to determine when the subset value iteration is more efficient than standard value iteration for a POMDP, it does not answer the question that how much savings the algorithm can bring about, which turns out to be a very difficult problem in theoretical analysis. The difficulty lies in not only the size of the set $\tau(\mathcal{B})$ but also the vectors representing the step and the optimal value functions. Let us assume that $\tau(\mathcal{B})$ is a proper subset of the belief space. We can imagine at least two cases. In one case, if at each iteration the step value function has very few useful vectors in the subset $\tau(\mathcal{B})$, the subset value iteration can be very efficient. In the other case, if at each iteration the step value function has all useful vectors in the subset $\tau(\mathcal{B})$, subset value iteration has the same complexity as standard value iteration in an asymptotic sense. In general, given a POMDP, it is difficult to predict how these vectors scatter around the belief subsets and the belief space. Consequently, it is hard to predict how much saving the subset value iteration algorithm can bring about for a POMDP without solving it.

## 4.5 Empirical Studies

We present our empirical results on two variants of a designed maze problem and the problems in a standard test-bed in this subsection. Some common settings to all experiments in the paper are as follows. The experiments are conducted on an UltraSparc II machine with dual CPUs and 256MB of RAM. Our codes are written in C and executed under a UNIX operating system Sola 2.6. When solving linear programs, we use a commercial package CPLEX V6.0. The discount factor is set at 0.95 and round-off precision is set at $10^{-6}$. When not stated otherwise, the quality requirement $\epsilon$ is set to 0.01. We use incremental pruning to compute representing sets of value functions over the belief space or belief subsets.

We compare the performances of subset and standard value iteration. For simplicity, we denote them respectively by `ssVI` and `VI`. At each iteration, we compare `VI` and `ssVI` in two measures: sizes of sets representing value functions and total time of DP updates.

### 4.5.1 The Maze Problem

The maze problem is specified in Figure 3. There are 10 locations and the goal is location 9. A robot agent can execute four "move" actions to change its position, optionally a "look" action to observe its surroundings and a "declare" action to announce its success of goal-attainment. The "move" actions can achieve their intended effects with probability 0.8, but might have no effects with probability 0.1 (the agent's position remains unchanged) or lead to overshooting with probability 0.1. Moving against maze walls leaves the agent at its original location. Other actions do not change the agent's position. At each time point, the robot receives a "null" observation giving no useful information at all, or reads four sensors so as to reason about its current position. Each sensor informs the robot whether there is a wall or nothing along a direction. In the figure thick lines stand for walls and thin lines for nothing (open). For instance, if the agent is at location 2, ideally a string "owow" (in the order of East, South, West and North) is received. Specific parameters will be instantiated in relevant empirical analysis. The robot is required to maximize the infinite discounted sum of rewards.

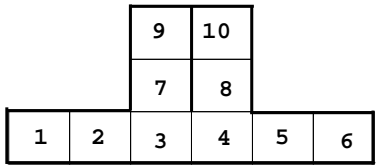

Figure 3: A maze problem

Two variants of the above maze are designed to test ssVI and VI. They are denoted by maze1 and maze2. For maze1, ssVI is more efficient; for maze2, ssVI is less efficient.

**Case I:** $\tau(\mathcal{B}) \subset \mathcal{B}$

The maze1 problem has a state space of 10 locations, an action space of size 5 (four "move" and one declare) and an observation space of size 6 (strings of four letters). An ideal string is received with certainty after any action is performed. When the agent declares goal at location 9, it receives a reward of 1 unit; if it does so at location 10, it receives a reward of −1. Other combinations of actions and observations lead to no reward.

We collect the results in Figure 4. The first chart in the figure depicts the total time of DP updates in log-scale for VI with the loose stopping criterion and ssVI with the strict one (Section 4.3.2). To compute a 0.01-optimal value function, VI took 20,000 seconds after 162 iterations while ssVI with strict stopping criterion took 900 seconds after 197 iterations. We note that ssVI needs more iterations but it still takes much less time. The performance difference is big. Moreover, more iterations means that the value function generated by ssVI is closer to optimality.

This is not a surprising result if we take a look at the matrix $P_{az}$ for an action $a$ and observation $z$. We know that the matrix impacts the size of the simplex $\tau(\mathcal{B}, a, z)$. The dimension of the matrix is $10 \times 10$. The entry of $P_{az}$ at $(i, j)$ is the product of the transition probability $P(s_j | s_i, a)$ and observation probability $P(z | s_j, a)$. Let us assume that

139

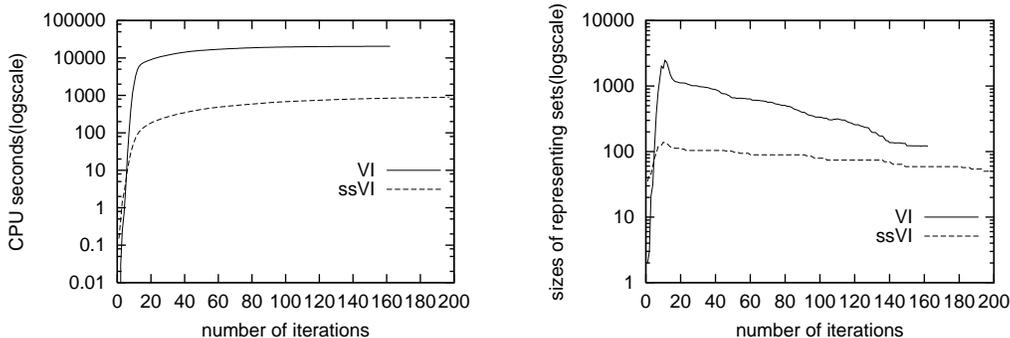

Figure 4: Comparative studies for `VI` and `ssVI` on Maze1

the observation be `owow`. Hence, the possible locations may be 2 or 5. Regardless of actions executed, only entries in row 2 and 5 of $P_{az}$ can be non-zero. Therefore, the matrix is highly sparse and non-invertible and the simplex $\tau(\mathcal{B}, a, z)$ is much smaller than $\mathcal{B}$. This analysis holds similarly for other combinations of actions and observations. Hence, `ssVI` accounts for only a small portion of the belief space. This explains why `ssVI` is more efficient than `VI`. In addition, we expect that the sets generated by `ssVI` are much smaller than those by `VI`.

This is confirmed in the second chart of the figure. It depicts the sizes of the sets representing the value functions generated by `ssVI` and `VI` at each iteration. When counting the size of $\mathcal{V}_n^{\tau(\mathcal{B})}$, we collect the sum of the sizes of representing sets over $|\mathcal{A}||\mathcal{Z}|$ simplices. We note that at the same iteration `VI` always generates much more vectors than `ssVI`. The sizes at both curves increase sharply at first iterations and then stabilize. The size for `VI` reaches its peak of 2466 at iteration 11 and the maximum size for `ssVI` is 139 at iteration 10. This size in `VI` is about 20 times many as that in `ssVI`. This is a magnitude consistent with the performance difference. After the sizes stabilize, the sizes of the sets generated by `VI` are around 130 and they are around 50 in `ssVI`.

**Case II:** $\tau(\mathcal{B}) = \mathcal{B}$

The problem `maze2` is designed to show that `ssVI` could be less efficient than `VI` when the selected belief subset $\tau(\mathcal{B})$ is equal to the belief space $\mathcal{B}$. The problem has a state space of 10 locations, an action space of size 6 (four moves, one `stay` and one `declare`) and an observation space of size 7 (6 strings and a `null` telling nothing). The action `stay` does not change the agent's position. `maze2` has more complications on the observation model. Due to hardware limitations, after a move action, with a probability of 0.1, the agent receives a wrong report where the string `owow` is collected as `owww` and `woww` as `wowo`. If the `declare` action is executed, the agent always receives a `null` observation. In addition, if the agent executes `stay`, it receives either a `null` observation with probability 0.9 or the ideal string about the surrounding locations with probability 0.1.

The reward model is accordingly changed to reflect new design considerations. We assume that the agent needs to pay for its information about states. For this purpose, if the agent executes `stay`, it really does nothing and thus yields no cost (i.e., *negative reward*). In contrast, the "move" actions always cause a cost of 2. Depending on the locations at which it executes `declare`, it receives rewards or costs: if the location is state 9, it receives a reward of 1; if state 10, it receives a cost of 1; otherwise, it leads to no rewards. The `stay` action is attractive in that it yields no cost but it leads to an useful observation about states with a small likelihood.

The empirical results are collected in Figure 5. First, we note both `VI` and `ssVI` are able to run only 11 iterations within a reasonable time limit (8 hours). The first chart in the figure presents the time costs along iterations. To run 11 iterations, `ssVI` takes 53,000 seconds while `VI` takes around 30,900 seconds. Therefore, `ssVI` is slower than `VI` for this problem. However, the magnitude of performance difference is not big. To explain this, let us consider the matrix $P_{az}$ for action `stay` and observation `null`. The transition matrix is an identity and each state can lead to the `null` observation with probability of 0.9 if `stay` is executed. Therefore, the matrix $P_{az}$ is invertible and the simplex $\tau(\mathcal{B}, a, z)$ is the same as the belief space $\mathcal{B}$. Because `ssVI` needs to account for additional simplices for other combinations of actions and observations, `ssVI` must be less efficient than `VI`. This explains the performance difference in time between `ssVI` and `VI`.

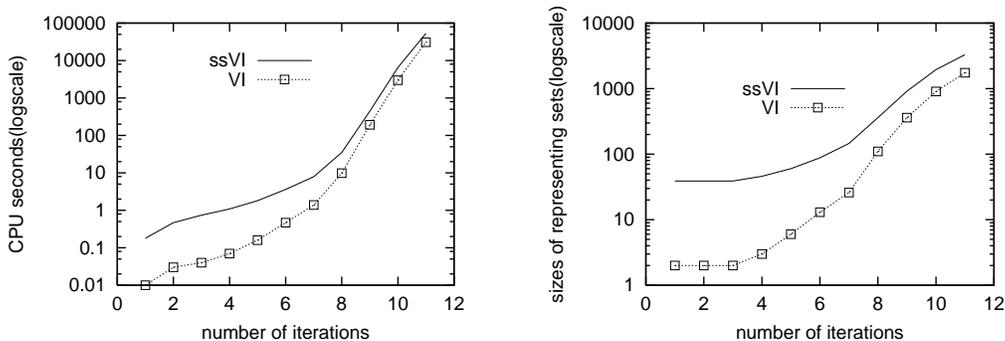

Figure 5: Comparative studies for `VI` and `ssVI` on maze2

It is also anticipated that `ssVI` should generate more vectors than `VI` at the same iteration because the size of $\mathcal{V}_n^{\tau(\mathcal{B})}$ is defined to be sum of the individual sets in it. This is confirmed and demonstrated in the second chart of Figure 5. The curve for `ssVI` is always on the upper side of that for `VI`. For the 11th iteration, `ssVI` generates 3,300 vectors and `VI` generates around 1,700 vectors.

### 4.5.2 MORE EXPERIMENTS ON THE TEST-BED

To validate the performance of the subset value iteration over different problem domains, we collected the results of the algorithm on the standard test-bed maintained by Tony Cassandra [6]. In the literature, the eight problems are commonly referred to as 4x3CO, Cheese, 4x4, Part Painting, Tiger, Shuttle, Network, and Aircraft. Table 2 presents detailed statistics for these problems. In the table, Rows 2–4 give the sizes of problem parameters, namely the number of states, observations and actions. Row 5 and 6 show the CPU seconds for the standard and subset value iteration algorithms to compute the 0.01-optimal policy for each problem. Row 7 shows the number of the vectors representing the 0.01-optimal value function in standard value iteration. In our experiments, we implemented the subset value iteration in the individual fashion. In Row 8, an entry takes the form $\cdot/\cdot$, denoting the total number of vectors over all $|\mathcal{A}| \cdot |\mathcal{Z}|$ simplices and the maximum number of vectors among these simplices when the subset value iteration terminates. The last row shows whether the belief subset $\tau(\mathcal{B})$ is a proper subset of the belief space.

---

6. See the URL `http://pomdp.org/pomdp/examples/index.shtml`

|  | 4x3CO | Cheese | 4x4 | Paint | Tiger | Shuttle | Network | Aircraft |
|---|---|---|---|---|---|---|---|---|
| $|\mathcal{S}|$ | 11 | 11 | 16 | 4 | 2 | 8 | 7 | 12 |
| $|\mathcal{Z}|$ | 4 | 4 | 2 | 4 | 2 | 2 | 2 | 5 |
| $|\mathcal{A}|$ | 11 | 7 | 4 | 2 | 3 | 3 | 4 | 6 |
| VI(Time) | 3.52 | 14.06 | 27.47 | 38.75 | 82.40 | 6130.69 | 13283.15 | 1723193.34 |
| ssVI(Time) | 63.28 | 85.44 | 85.44 | 85.20 | 145.21 | 1437.32 | 2810.21 | 425786.49 |
| VI(#) | 4 | 14 | 20 | 9 | 9 | 208 | 491 | 2071 |
| ssVI(#) | 43/1 | 32/2 | 42/10 | 22/9 | 22/9 | 98/45 | 201/50 | 3236/428 |
| subspace | yes | yes | yes | no | no | yes | yes | yes |

Table 2: Comparative studies for `ssVI` and `VI` on the standard test-bed

In discussing the performances of the subset value iteration algorithm, we categorize the tested problems into three classes. In the first class, the subset $\tau(\mathcal{B})$ is actually the same as the belief space. Subset value iteration must be less efficient than standard value iteration. The reason is as follows: there exists at least one belief simplex such that value iteration over it has the same complexity as standard value iteration; moreover, subset value iteration needs to account for other simplices. Example problems are tiger and paint. Let us take the paint problem as an instance. Our results show that there are two simplices are the same as the belief space. The 0.01-optimal value functions over them are represented by 9 vectors, each of which has the same number of vectors representing the 0.01-optimal value function over the entire belief space. In the second class of the tested problems, the set $\tau(\mathcal{B})$ is a proper subset of the belief space, and meanwhile the numbers of the vectors representing value functions over the belief space and individual simplices are very small. Subset value iteration may not be so efficient as standard value iteration because of the overhead of accounting for a large number of simplices. Example problems include 4X3CO, cheese and 4X4. Let us take 4X3CO as an instance. The 0.01-optimal value function over the entire belief space is represented by only 4 vectors, whereas the 0.01-optimal value function over each simplex is represented by only 1 vector. Since the subset value iteration has to account for 44 simplices, subset value iteration is less efficient than standard value iteration. In the third class of tested problems, the set $\tau(\mathcal{B})$ is a proper subset of the belief space, and meanwhile the numbers of the vectors representing value functions over the belief space are moderately large. The subset value iteration algorithm is more efficient than the standard value iteration algorithm. Examples include shuttle, network and aircraft. Let us take network as an instance. The 0.01-optimal value function in the belief space is represented by 491 vectors, whereas the 0.01-optimal value function in $\tau(\mathcal{B})$ is represented by less than 201 vectors (note that there are duplicates across belief simplices). The maximum size of the representing sets over all simplices is 50. In this case, we expect that the savings brought by the subset value iteration outweighs the overhead of accounting for the simplices. The result shows that subset value iteration is about 5 times faster than standard value iteration.

Combining these results with those on the maze problem, we see that the computational savings brought by the subset value iteration vary with different problem domains. Theorem 3 can be used to determine whether subset value iteration can bring about computational savings for a POMDP. In the event that the belief set $\tau(\mathcal{B})$ is a proper subset of the belief space, the magnitude of the savings needs to be determined through empirical evaluation.

## 5. Informative POMDPs

In this section, we study a special POMDP class, namely informative POMDPs. For this POMDP class, there are natural belief subsets for value iteration to work with. We will show how to formally define these subsets. As the value iteration over these belief subsets has been described (Zhang & Liu, 1997), our focus is to compare the algorithm with the general subset value iteration developed in the previous section.

### 5.1 Motivation

As noted by some authors, in reality an agent often has a good, although imperfect, idea about its locations (Roy & Gordon, 2002). For instance, mobile robots and other real world systems have local uncertainty, but rarely encounter global uncertainty. Let us exemplify this using the maze in Figure 3. Suppose that at each time point an agent receives a string of four letters with certainty. In total, there are 6 observations, `owww`, `owow`, `owoo`, `wwow`, `wowo` and `woww` regardless of executed actions. If we enumerate all possible observations and the set of locations at which the agent receives such observations, we end up with the following table.

| observations | states | observations | states |
|:---:|:---:|:---:|:---:|
| `owww` | {1 } | `owow` | { 2, 5} |
| `owoo` | { 3,4 } | `wwow` | { 6 } |
| `wowo` | { 7,8} | `woww` | { 9,10 } |

On the other hand, the strings can be used to infer the agent's locations. For instance, if a string `owoo` is received, the world must be at location 3 or 4. Hence, the observation `owoo` restricts the world into a small range of world states. In fact, any observation can restrict the world into at most two states although the world has ten. For this reason, the POMDP is said to be *informative*.

In general, an agent perceives the world via observations. Starting from any state, if the agent executes an action $a$ and receives an observation $z$, the world states can be categorized into two classes by the observation model: states the agent can be in and states it cannot. Formally, the former is $\{s | s \in \mathcal{S} \text{ and } P(z|s,a) > 0\}$. The set is denoted by $\mathcal{S}^{az}$. We use the set to define the informativeness. An $[a, z]$ pair is said to be *informative* if the size $|\mathcal{S}^{az}|$ is much smaller than $|\mathcal{S}|$. An observation $z$ is *informative* if $[a, z]$ is informative for every action $a$ giving rise to $z$. A POMDP is *informative* if all observations are informative.

In informative POMDPs, since any observation restricts the world into a small set of states, the agent knows that the world cannot be in the state outside this small set. In other words, for those states outside the set, the agent has zero beliefs. Consequently, an observation can also restrict belief states into a belief subset.

## 5.2 Belief Subset Selection

For informative POMDPs, we can select belief subset $\tau(\mathcal{B})$ as before. Combining the informativeness assumption and Corollary 1, we know that $\tau(\mathcal{B})$ is a proper subset of the belief space. So, value iteration over $\tau(\mathcal{B})$ carries the space and time savings. In this section, we choose an alternative belief subset for value iteration. Compared against the subset $\tau(\mathcal{B})$, the subset that we choose yields several advantages. First, it is conceptually simple and geometrically intuitive. Second, it facilitates employing the low dimensional representation of vectors. Third, it may lead to additional savings in time if the observation models of a POMDP are independent of actions. The latter two advantages will be shown later.

To define the belief subset (say $\phi(\mathcal{B})$), we first define a subset $\phi(\mathcal{B}, a, z)$ for an action and observation pair. Then, the belief subset $\phi(\mathcal{B})$ is formed by taking the union of $\phi(\mathcal{B}, a, z)$ over all action and observation pairs. To be specific,

$$\phi(\mathcal{B}, a, z) = \{b| \sum_{s \in \mathcal{S}^{az}} b(s) = 1.0, \ \forall s \in \mathcal{S}^{az}, b(s) \geq 0\} \tag{12}$$

and

$$\phi(\mathcal{B}) = \cup_{a,z} \phi(\mathcal{B}, a, z).$$

It is trivial to see that $\phi(\mathcal{B}, a, z)$ is a belief simplex. It can be proven that for any belief state $b$, $\tau(b, a, z)$ must be in $\phi(\mathcal{B}, a, z)$. Therefore, $\tau(\mathcal{B}, a, z)$ is a subset of $\phi(\mathcal{B}, a, z)$. Consequently, $\tau(\mathcal{B})$ is a subset of $\phi(\mathcal{B})$. This is summarized in the lemma below. The lemma is useful when we discuss the value iteration algorithm working with the belief subset $\phi(\mathcal{B})$.

**Lemma 5** *For a POMDP, $\tau(\mathcal{B}) \subseteq \phi(\mathcal{B})$.*

It is of interest to compare the $\tau$-simplex and $\phi$-simplex for a pair of $a$ and $z$. Although both simplices are generated by a list of belief states, $\phi$-simplex has more intuitive geometric meaning. Each belief state in the basis of $\phi(\mathcal{B}, a, z)$ is a unit vector, i.e., it has probability mass on one state. Therefore, the belief state in the basis must be a boundary point of the belief space. In contrast, a belief state in the basis of a $\tau$-simplex can be an interior point. See Figure 1 for an example, where $A_2$, $A_3$ are interior points and $A_1$ is a boundary point of the belief space.

## 5.3 Value Iteration over $\phi(\mathcal{B})$

From the theoretical perspective, the feasibility of conducting value iteration in $\phi(\mathcal{B})$ is justified by Lemma 1. Combined with Lemma 5, the subset $\phi(\mathcal{B})$ is a closed set. Hence, the MDP theory is applicable to defining the DP update equation. By our discussions on the relationship and value iteration in Section 4.3, value iteration working with $\phi(\mathcal{B})$ retains the quality of value functions.

We further exploit the informative feature in value iteration over $\phi(\mathcal{B})$. We briefly outline the subset value iteration algorithm and refer the readers to a detailed description (Zhang

& Liu, 1997). The basic idea is to reduce the dimensions of vectors in representing sets of value functions. Note that for any pair $[a, z]$, since the beliefs in states outside $\mathcal{S}^{az}$ are zero, a vector in the representing set of a value function over the simplex $\phi(\mathcal{B}, a, z)$ needs only $|\mathcal{S}^{az}|$ components. In an individual fashion, a DP update over $\phi(\mathcal{B})$ computes a collection $\{\mathcal{V}_{n+1}^{\phi(\mathcal{B}, a, z)}\}$ from a collection $\{\mathcal{V}_n^{\phi(\mathcal{B}, a, z)}\}$ where $\mathcal{V}_n^{\phi(\mathcal{B}, a, z)}$ is the $n$th-step value function and the vectors in it have $|\mathcal{S}^{az}|$ dimensions. The procedure of conducting a DP update is parallel to that in Section 4.2 except that $\tau(\mathcal{B}, a, z)$ is replaced by $\phi(\mathcal{B}, a, z)$. In the enumeration step, when building a vector $\beta$ in a belief simplex $\phi(\mathcal{B}, a, z')$ using Equation (10), we need only define its components corresponding to the set $\mathcal{S}^{a'z'}$. In the reduction step, for each constructed set $\mathcal{V}_{n+1}^{\phi(\mathcal{B}, a', z')}$, a pruning procedure is called to remove useless vectors to obtain the minimal representation of the set. Note that the lower dimension feature is also used to cut down the number of variables in setting up linear programs.

Interestingly, DP updates over $\phi(\mathcal{B})$ account for a larger subset than those over $\tau(\mathcal{B})$. Hopefully, since DP updates over $\phi(\mathcal{B})$ explicitly employ the economy of representation, they could be more efficient. In addition, DP updates over $\phi(\mathcal{B})$ have another advantage in the event that the observation models of a POMDP are independent of actions, i.e., the probabilities $P(z|s, a)$ being independent of $a$. Hence, given an observation $z$, the simplices $\phi(\mathcal{B}, a, z)$ are the same for all actions. Therefore, DP updates over $\phi(\mathcal{B})$ only account for $|\mathcal{Z}|$ $\phi$-simplices. However, DP updates over $\tau(\mathcal{B})$ usually need to account for $|\mathcal{A}||\mathcal{Z}|$ $\tau$-simplices because an observation determines different $\tau$-simplices when combined with different actions.

### 5.4 Empirical Studies

We have conducted experiments to compare `VI`, `ssVI` and `infoVI`, which refers to value iteration exploiting the low-dimension feature. The experiments on `maze1` (defined in Section 4.5) can be found elsewhere (Zhang & Zhang, 2001; Zhang, 2001). The results, together with existing results (Zhang & Liu, 1997), showed that value iteration over $\phi(\mathcal{B})$ can be significantly more efficient than standard value iteration. For reference, we mention that it is feasible to integrate a point-based technique and value iteration over $\phi(\mathcal{B})$ in order to take advantage of both reducing the iteration number and accelerating the iterative steps (Zhang & Zhang, 2001b). To demonstrate this, we include results on a 96-state POMDP in Appendix B.

### 5.5 Restricted Value Iteration and Dimension Reduction

We compare the value iteration algorithms in this and the previous section. Through the comparison, we would like to emphasize that working with belief subsets does not imply working with low-dimensional vectors.

Although both algorithms work with belief subsets, the mechanisms exploited to achieve the computational gains are different. The general value iteration works with the belief subset $\tau(\mathcal{B})$ but the dimension of representing vectors is the same as the number of states, whereas value iteration over $\phi(\mathcal{B})$ works with a superset of $\tau(\mathcal{B})$ but the dimension of the vectors is smaller than the number of the states. To facilitate demonstrating how a reduced belief set and the low-dimensional representation respectively contribute to the computational gains, we experimented with a carefully designed maze problem that is amenable to both algorithms. However, it is worth pointing out that working with a reduced belief set does not mean that the vectors can be represented in low dimensions. We illustrate this point by continuing our discussions of the example in Section 3.3. Such an example shows that the primary advantage of value iteration over $\tau(\mathcal{B})$ stems from the size of the chosen belief subset rather than dimension reduction of the representing vectors.

**Example (Continued)** For the POMDP example presented in Section 3.2, the subset $\tau(\mathcal{B})$ consists of only two line segments in the entire belief space. Clearly value iteration over $\tau(\mathcal{B})$ is more efficient than standard value iteration. However, if one runs value iteration over $\phi(\mathcal{B})$ on this POMDP anyhow, the algorithm is less efficient than the standard value iteration algorithm. This follows from (1) the set $\mathcal{S}^{a_i z_j}$ is equal to the set of states $\mathcal{S}$ for any action $a_i$ and observation $z_j$ by the second assumption, and (2) each $\phi$-simplex is actually the same as the belief space by the definition in Equation (12). To solve this POMDP, the susbet value iteration algorithm is definitely a better choice than the value iteration algorithm for informative POMDPs. □

## 6. Near-Discernible POMDPs

In this section, we study near-discernible POMDPs. For this POMDP class, we develop an anytime value iteration algorithm working with growing belief subsets.

### 6.1 Motivation

A discernible POMDP assumes that once in a while the uncertainty about world states vanishes if a particular action is executed and the observations pertain to the action fully reveal the identities of the world (Hansen, 1998). Our research on near-discernible POMDPs was motivated by two aspects. One of them arises from the origin of applying POMDP as a framework for planning under uncertainty. To achieve a goal location, an agent has to not only change its positions by performing goal-achieving actions but also reason about its surroundings by performing information-gathering actions. However, at one time point the agent cannot simultaneously move its positions and observe its environments. For instance, if an information-gathering action is performed, the agent cannot move its positions meanwhile. The other aspect motivating the concept of near-discernibility arises from existing research in the community. Near-discernible POMDPs generalize discernible POMDPs in that even when an information-gathering action is performed, the agent can get a rough, rather than exact, idea about world states and uncertainty vanishes in some sense.

We revise the maze problem to fix ideas on the first motivation. The action space consists of six actions: four "moving" actions, `look` and `declare`. If move actions or `declare` are performed, an observation `null` is received and the agent gets no information at all. If `look` is performed, an ideal string is received and the agent gets imperfect information since different locations might yield the same string. On one hand, to achieve the goal location, the agent has to change its positions. On the other hand, to declare goal attainment with confidence, it has to perform `look` and reason about the environment. Arbitrarily declaring goal attainment leads to a penalty. Consequently, at a time point the agent faces the problem of choosing a move or `look`.

We note that the subset value iteration algorithm usually yields no computational advantage for near-discernible POMDPs. We give an example in which the subset $\tau(\mathcal{B})$ is the same as the belief space $\mathcal{B}$ under some assumptions. Suppose the maze is a square grid. Locations are numbered such that at each row the indices of the locations increase from left to right. We assume that a move action achieves its intended effects with a high likelihood, but may have no effect (i.e., the agent's location remains unchanged) or may lead to overshooting with a small probability. Under these assumptions, the transition matrix for action `east` is upper-triangular and invertible. If at each location `null` is received with a positive probability after a move, the transformational matrix $P_{\texttt{east,null}}$ is invertible. By Theorem 3, the belief subset $\tau(\mathcal{B},\texttt{east},\texttt{null})$ is equal to the belief space $\mathcal{B}$.

Our solution to near-discernible POMDPs rests on the intuition that the agent needs to interleave goal-achieving actions and information-gathering actions. A typical sequence of executed actions should consist of several goal-achieving actions and an information-gathering action. The difficulty is how frequently the agent should execute an information-gathering action. In this section, we consider the action and observation sequences containing more goal-achieving actions incrementally. We show that such sequences can be used to determine belief simplices. As more sequences are added, the union of the belief simplices grows. In the following, we give some technical preparations and then describe the algorithm designed for near-discernible POMDPs. In order to put our discussions under a general context, we shall use *information-rich* and *information-poor* actions instead of information-gathering and goal-achieving actions respectively.

## 6.2 Histories, Belief Subsets and Value Functions

A history is a sequence of ordered pairs of actions and observations. We usually denote a history by $h$. The number of pairs of actions and observations is referred to as *the length of a history*. A history of length $l$ is denoted by $[a_1, z_1, \cdots, a_l, z_l]$. If an agent's initial belief state is $b$ and a history $h$ of length $l$ is realized, its belief state can be updated at each time step. The notation $\tau(b, h)$ denotes the belief at the time point $l$. The set $\tau(\mathcal{B}, h)$ is defined to be $\cup_{b \in \mathcal{B}} \tau(b, h)$, consisting of all possible belief states that the agent can be in at step $l$ if it starts with any belief and history $h$ is realized. Note that if $h$ is of length 1(say $h = [a, z]$), $\tau(\mathcal{B}, h)$ degenerates to our previous notation $\tau(\mathcal{B}, a, z)$.

**Lemma 6** *For any history $h$, the belief subset $\tau(\mathcal{B}, h)$ is a simplex.*

A set of histories is usually denoted by $\mathcal{H}$. The belief subset $\tau(\mathcal{B}, \mathcal{H})$ denotes the union of simplices for all histories in the set $\mathcal{H}$, i.e., $\cup_{h \in \mathcal{H}} \tau(\mathcal{B}, h)$. Value functions over the simplex $\tau(\mathcal{B}, h)$ and belief subset $\tau(\mathcal{B}, \mathcal{H})$ are referred to as $V^{\tau(\mathcal{B}, h)}$ and $V^{\tau(\mathcal{B}, \mathcal{H})}$ respectively. Given a set $\mathcal{V}^{\tau(\mathcal{B}, h)}$ representing value function $V^{\tau(\mathcal{B}, h)}$, the procedure $\texttt{simplexPrune}(\mathcal{V}, B_{\tau(\mathcal{B}, h)})$ in Table 1 computes the minimal representation of $\mathcal{V}^{\tau(\mathcal{B}, h)}$. In the context of history, the occurrences of the basis $B_{\tau(\mathcal{B}, a, z)}$ should be replaced by $B_{\tau(\mathcal{B}, h)}$.

## 6.3 Space Progressive Value Iteration

We describe the space progressive value iteration (SPVI) algorithm. As an anytime algorithm, SPVI begins with a belief subset and gradually grows it. When a certain stopping criterion is met, SPVI terminates and returns a set of vectors for the agent's decision making.

### 6.3.1 ALGORITHMIC STRUCTURE

SPVI interleaves value iteration (computing a value function for a belief subset) and subset expansion (expanding the current belief subset to a larger one). The belief subsets in SPVI are introduced by sets of histories. Subset expansion is achieved through incorporating more histories. For convenience, the set of histories determining the $i$-th belief subset is denoted by $\mathcal{H}_i$. The belief subset determined by $\mathcal{H}_i$ is $\tau(\mathcal{B}, \mathcal{H}_i)$. The value function constructed by SPVI for $\mathcal{H}_i$ is $\mathcal{V}^{\tau(\mathcal{B}, \mathcal{H}_i)}$.

The pseudo-code in Table 3 implements SPVI. A set of histories $\mathcal{H}_0$ (and therefore the belief subset $\tau(\mathcal{B}, \mathcal{H}_0)$), a value function $\mathcal{V}^{\tau(\mathcal{B}, \mathcal{H}_0)}$ and the quality precision $\eta$ are initialized at line 1. This step can be regarded as the 0th-step expansion of belief subset. Note that we set the initial value function to be the minimum reward for all pairs of actions and states. (This is for the convergence issue discussed later.) Value iteration over the current subset $\tau(\mathcal{B}, \mathcal{H}_i)$ is conducted at line 3, and the belief subset is expanded to the subset $\tau(\mathcal{B}, \mathcal{H}_{i+1})$ through constructing a superset $\mathcal{H}_{i+1}$ of the current set $\mathcal{H}_i$ at line 4. Value function $\mathcal{V}^{\tau(\mathcal{B}, \mathcal{H}_i)}$ for the current belief subset is set to be the initial value function for the next subset at line 5. If the stopping condition is not satisfied at line 7, SPVI goes to the next iteration; otherwise, it terminates and returns the latest value function $\mathcal{V}^{\tau(\mathcal{B}, \mathcal{H}_{i-1})}$.

To ensure the efficiency of SPVI, its initial belief subset should be chosen to be small. To this end, we set $\mathcal{H}_0$ to be $\{[a, z] \mid a \in \mathcal{A}_{IR}, z \in \mathcal{Z}_{IR}\}$ where $\mathcal{A}_{IR}$ is the set of information-rich actions and $\mathcal{Z}_{IR}$ is the set of observations led to by those actions. The subset $\tau(\mathcal{B}, \mathcal{H}_0)$ is small due to the discernability property.

In the sequel, we discuss value iteration in a belief subset, subset expansion and the stopping criterion in detail.

### 6.3.2 VALUE ITERATION IN A BELIEF SUBSET

Given a set $\mathcal{V}$ of vectors, a set $\mathcal{H}$ of histories and a precision threshold $\eta$, value iteration computes an improved value function over the belief subset $\tau(\mathcal{B}, \mathcal{H})$. This is accomplished by conducting a sequence of DP updates. In the following, we discuss implicit DP updates, the convergence issue and the stopping criterion in the value iteration step.

An implicit DP update computes a new value function from the current one for belief subset $\tau(\mathcal{B}, \mathcal{H})$. Let $\mathcal{U}_j$ ($\mathcal{U}_0 = \mathcal{V}$) denote the $j$-step value function. Thus, a DP update computes value function $\mathcal{U}_{j+1}$ from $\mathcal{U}_j$. The procedure of computing $\mathcal{U}_{j+1}$ from $\mathcal{U}_j$ is parallel to the collective DP update in Section 4. In particular, when defining a vector $\beta_{a,\delta}$ given an action $a$ and a mapping $\delta$ in Equation (9), the occurrences of $\mathcal{V}_n^{\tau(\mathcal{B})}$ are replaced by $\mathcal{U}_j$. By enumerating actions and mappings, all defined vectors form the set $\mathcal{U}_{j+1}$. Its minimal representation is obtained by removing useless vectors w.r.t. the subset $\tau(\mathcal{B}, \mathcal{H})$.

The convergence issue arises because the subset $\tau(\mathcal{B}, \mathcal{H})$ may not be a closed set. To guarantee the convergence of value iteration, we set $\mathcal{U}_{j+1}$ to be the union of set $\mathcal{U}_{j+1}$ and $\mathcal{U}_j$ after a DP update. Together with the fact that the initial value function is set to be the minimum reward for all actions and states, the sequence $\{\mathcal{U}_j\}$ monotonically increases in terms of induced value functions. On the other hand, the value functions in the

```
SPVI:
1.  i ← 0, initialize H_0, V^{τ(B,H_0)} ← min_{s∈S,a∈A} r(s,a), η ← ε(1 − λ)/2λ
2.  Do
3.      V^{τ(B,H_i)} ← subsetVI(V^{τ(B,H_i)}, H_i, η)
4.      < H_{i+1}, τ(B, H_{i+1}) >← expandSubset(V^{τ(B,H_i)}, H_i)
5.      V^{τ(B,H_{i+1})} ← V^{τ(B,H_i)}
6.      i ← i + 1
7.  Until (stopping condition is met)
8.  Return V^{τ(B,H_{i−1})}

subsetVI(V, H, η):
1.  j ← 0, U_0 ← V
2.  Do
3.      U_{j+1} ← subsetDPUpdate(U_j, τ(B, H))
4.      U_{j+1} ← U_{j+1} ∪ U_j
5.      j ← j + 1
6.  While ( max_{b∈τ(B,H)} |U_j(b) − U_{j−1}(b)| ≤ η )
7.  Return U_{j−1}

expandSubset(V, H)
1.  H' ← H
2.  For each β in the set V
3.      If β.history is maximal in H and  β.action is information-poor
4.          For each [a, z] in A_{IP} × Z_{IP}
5.              H' ← H' ∪ {[h, a, z]}
6.  Return < H', τ(B, H') >
```

Table 3: Space progressive value iteration (SPVI)

sequence are upper bounded by the optimal value function. Consequently, value iteration in $\tau(\mathcal{B}, \mathcal{H})$ must converge. As a result, the Bellman residual between value functions, $\max_{b \in \tau(\mathcal{B},\mathcal{H})} |\mathcal{U}_{j+1}(b) - \mathcal{U}_j(b)|$, becomes smaller in $\tau(\mathcal{B}, \mathcal{H})$ as value iteration continues. When the residual falls below the threshold $\eta$, value iteration terminates.

The value iteration step is implemented as the procedure subsetVI in Table 3. Given a set $\mathcal{V}$ of vectors, a set $\mathcal{H}$ of histories and a threshold $\eta$, the procedure computes an improved value function for belief subset $\tau(\mathcal{B}, \mathcal{H})$. Value function $\mathcal{U}_0$ is set to be the input set $\mathcal{V}$ at line 1. The new value function $\mathcal{U}_{j+1}$ is computed by a DP update at line 3. To guarantee convergence, $\mathcal{U}_{j+1}$ to set to be the union of $\mathcal{U}_j$ and $\mathcal{U}_{j+1}$ at line 4. The stopping criterion is tested at line 6. If it is met, the latest value function $\mathcal{U}_{j−1}$ is returned.

### 6.3.3 Subset expansion

Given a set $\mathcal{V}$ of vectors and a set $\mathcal{H}$ of histories, the subset expansion step expands the belief subset $\tau(\mathcal{B}, \mathcal{H})$ to a larger one. This is achieved by generating a superset $\mathcal{H}'$ of $\mathcal{H}$. The new belief subset $\tau(\mathcal{B}, \mathcal{H}')$ is thus a superset of $\tau(\mathcal{B}, \mathcal{H})$. Hence, the key in subset expansion is how to generate a history set $\mathcal{H}'$. In the following, we propose two approaches to generating the history set using our intuition for near-discernible POMDPs. Both approaches generate new histories by exploiting the vectors in $\mathcal{V}$. We begin with an analysis of the vectors in the set $\mathcal{V}$ and show how to use them to generate histories.

Let $\beta$ be a vector in the set $\mathcal{V}$. Remember that $\beta$ is defined by a pair of action $a$ and mapping $\delta$. For convenience, such an action is said to be the *associated action of $\beta$*. In addition, if $\beta$ is useful in the set $\mathcal{V}$ w.r.t. the belief subset $\tau(\mathcal{B}, \mathcal{H})$, there must exist a history $h$ in $\mathcal{H}$ such that $\beta$ is useful w.r.t. the belief simplex $\tau(\mathcal{B}, h)$. The history $h$ is said to be the *associated history of $\beta$*. The associated history of a vector can be used to generate new histories by *extending the history*, i.e., appending the pairs of informative-poor actions and observations to the history. Let $\mathcal{A}_{IP}$ be the set of the information-poor actions and $\mathcal{Z}_{IP}$ be the set of the observations led to by those actions. Extending history $h$ results in a set $\{[h, a, z] | a \in \mathcal{A}_{IP}, z \in \mathcal{Z}_{IP}\}$. The set contains $|\mathcal{A}_{IP}||\mathcal{Z}_{IP}|$ histories. Each history in the set is called an *extension of history $h$*.

To generate $\mathcal{H}'$ from the set $\mathcal{H}$ and the set $\mathcal{V}$, one generic approach works as follows. Each vector $\beta$ in the set $\mathcal{V}$ is examined. If its associated history is long and the associated action is information-poor, we produce all extensions of its associated history. An extension is added to $\mathcal{H}'$ if it is not in $\mathcal{H}'$. (The reason is that the associated action of a vector should be information-rich if its associated history is sufficiently long.) To ensure that $\mathcal{H}'$ is a superset of $\mathcal{H}$, we set $\mathcal{H}'$ to be $\mathcal{H}$ in the beginning. Apparently, this approach to generating histories suffers from the exponential increase of the size $|\mathcal{H}'|$ in $|\mathcal{A}_{IP}|$ and $|\mathcal{Z}_{IP}|$. In the worst case where all the vectors in $\mathcal{V}$ are associated with information-poor actions, the size $|\mathcal{H}'|$ is $|\mathcal{H}||\mathcal{A}_{IP}||\mathcal{Z}_{IP}|$. Consequently, after the $i$-step subset expansion, $|\mathcal{H}_{i+1}|$ can be as large as $|\mathcal{A}_{IR}||\mathcal{Z}_{IR}|(|\mathcal{A}_{IP}||\mathcal{Z}_{IP}|)^i$ where $|\mathcal{A}_{IR}||\mathcal{Z}_{IR}|$ is the size of initial history set.

To alleviate this problem, we use a heuristic approach to generating $\mathcal{H}'$ in hope that the size $\mathcal{H}'$ increases moderately. The above exhaustive approach extends the histories associated with the vectors prescribing informative-poor actions. The heuristic approach does not extend all such histories. Instead it extends only *maximal histories* in the set $\mathcal{H}$. (A history is said to be maximal in a set if none of its extensions is in the set.) This is the only change made from the above approach. As indicated in the experiments, this restriction can effectively cut down the size of the history sets. Nonetheless, the heuristic approach shares the same worst-case complexity with the exhaustive approach.

The subset expansion step is implemented as the procedure `subsetExpansion` in Table 3. Given a set $\mathcal{V}$ of vectors and a set $\mathcal{H}$ of histories, it computes an expanded set $\mathcal{H}'$ and an expanded belief subset $\tau(\mathcal{B}, \mathcal{H}')$. The set $\mathcal{H}'$ is initialized to be $\mathcal{H}$ at line 1. For each vector $\beta$ in $\mathcal{V}$ at line 2, if its associated history is maximal in $\mathcal{H}$ and its action is information-poor (line 3), all the extensions of its associated history are added to $\mathcal{H}'$ (line 5). The expanded set $\mathcal{H}'$ and also the expanded belief subset $\tau(\mathcal{B}, \mathcal{H}')$ are returned at line 6.

### 6.3.4 STOPPING CRITERION AND DECISION-MAKING

As an anytime algorithm, SPVI can be terminated if a hard deadline is reached. Another stopping criterion of interest can be set as follows. Given a sufficiently large amount of time, SPVI would account for as many histories as possible. If a (near) optimal policy of the POMDP requires that information-rich actions be executed after a sequence of information-poor actions, SPVI should be able to compute a value function for the belief subset, which consists of all possible belief states that the agent encounters if it is guided by such a policy. After sufficiently many expansions of history sets and hence belief subsets, any vector associated with a maximal history prescribes an information-rich action. If all the vectors in the representing set prescribe information-rich actions, SPVI terminates. If a (near) optimal policy has the desired structure in its sequence of actions, the output value function should be near optimal in the final belief subset.

When SPVI terminates, the value function $\mathcal{V}^{\tau(\mathcal{B}, \mathcal{H}_{i-1})}$ can be used for decision making. Similarly to Equation (3), a $\mathcal{V}^{\tau(\mathcal{B}, \mathcal{H}_{i-1})}$-improving policy can be defined over the belief space.

### 6.3.5 Efficiency of SPVI

The efficiency of SPVI depends on the selected belief subsets. If these belief subsets are close to the belief space in size, SPVI must be inefficient. Fortunately, our approach for belief subset expansion ensures that the initial belief subset is small and the subsequent subsets grow slowly. First, since $\mathcal{H}_0$ is the set of the pairs of information-rich actions and observations, the initial belief $\tau(\mathcal{B}, \mathcal{H}_0)$ is relatively small. Second, the subsequent belief subsets $\tau(\mathcal{B}, \mathcal{H}_i)$ do not grow too quickly. The reason follows. In extending a history, the information-poor pairs are added to its end. Hence, the first action and observation pair of the histories in a set $\mathcal{H}_i$ must be information-rich. Therefore, for a history $h$ in the set $\mathcal{H}_i$, $\tau(\mathcal{B}, h)$ is small in size. Meanwhile, due to the heuristic for generating history sets, the sizes $|\mathcal{H}_i|$ would not increase too fast. These characteristics make SPVI efficient when compared with the standard value iteration algorithm.

Although the above analysis is empirically confirmed in our experiments below, it is worthy to mention that in the worst case the number of belief simplices grows exponentially in the number of $|\mathcal{A}_{IP}||\mathcal{Z}_{IP}|$. Since a history determines a belief simplex, in the worst case the number of the belief simplices after the $i$-step subset expansion is the same as the number of histories, i.e., $|\mathcal{A}_{IR}||\mathcal{Z}_{IR}|(|\mathcal{A}_{IP}||\mathcal{Z}_{IP}|)^i$ (see the third paragraph of Section 6.3.3).

## 6.4 Empirical Results

Since SPVI works in an anytime manner, our primary interest is to demonstrate how the quality of the generated value functions varies with the time cost. However, the availability of optimal solutions strongly depends on the "tractability" of the problems. If a near optimality is available, we compare it directly with the value function generated by SPVI by simulations. Otherwise, we simply compare value functions from SPVI against those from an approximate algorithm QMDP (Littman et al., 1995; Hauskrecht, 2000). Although the comparison is not strict in a formal sense, it can provide clues about the quality of value functions.

We report our results on two variants of the base maze problem and an office navigation problem. In one variant, SPVI terminated after a finite number of iterations and the output value function is near optimal; in the other variant, SPVI can quickly find a high-quality value function as time goes by (Zhang & Zhang, 2001a; Zhang, 2001). In the rest of this section, we report our results on the office navigation problem.

The environment is modeled after the floor plan of the authors' home department. The layout is shown in Figure 6. There are 35 states: 34 locations plus one terminal state. The action space is of size 6 (four `move`, one `look` and one `beep` replacing `declare` in the maze problem). Any action except `look` leads to a `null` observation. To introduce other observations, we note that in the figure, black bars represents doors and grey bars represent walls with display boards. The `look` action yields observation of strings of four letters for a location indicating, for each of the four directions, where there is a door (d), an empty wall (w), a wall with a display board (b), or nothing (o). In total, there are 22 different strings. Hence, plus the `null` observation, the observation space is of size 23. Transition probabilities for moves are specified identically as in the maze problem. Neither `look` nor `beep` changes the states of the environment. At each location, `look` produces the ideal string for that location with probability 0.75. With probability 0.05, it produces the `null` observation. Also with probability 0.05, it produces a string that is ideal for some other location and that differs from the ideal string for the current location by only 1 character. The robot receives a reward of 50 when beeping at location 22 and a reward of -10 when beeping at any other location (we don't want the robot to make a lot of noise). The `move` actions bring about a reward of -2 if they lead the robot to bumping into walls or doors. They have no rewards otherwise. The reward for the `look` action is always -1. The robot needs to get to location 22 and beep so that someone in the main office can come out and hand the robot some mail.

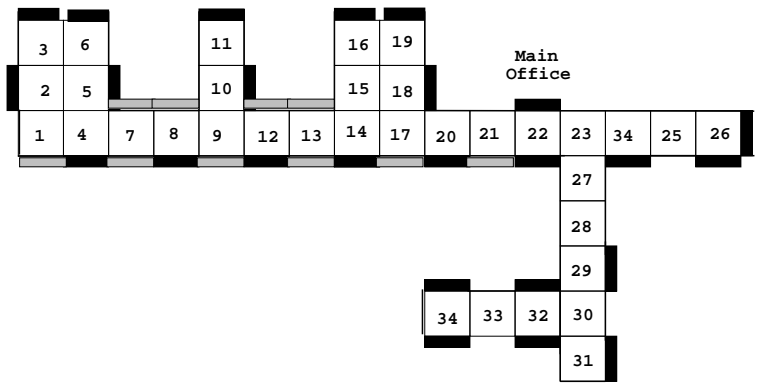

Figure 6: HKUST-CSD office environment

We conduct simulations about the generated value functions because no existing exact algorithm can find the near optimal value function. The simulation consists of 1000 trials. Each trial starts from a random initial belief state and is allowed to run up to 100 steps. The average reward across all the trials is used as a measurement of the quality of policies derived by value functions.

Figure 7 presents the results about the quality against time costs. We see that SPVI found a policy whose average reward is 19.6 in about 80,000 seconds. SPVI was manually terminated after running about 24 hours. It is found that the algorithm conducted three steps of subset expansion. By our data, after the first and second expansion steps, both rewards by simulation are 18.4. This is not far from 19.6 obtained after the third expansion step although it is difficult to say how close those polices are to the optimal. Compared with the solutions generated by QMDP, the policies generated by SPVI are clearly better.

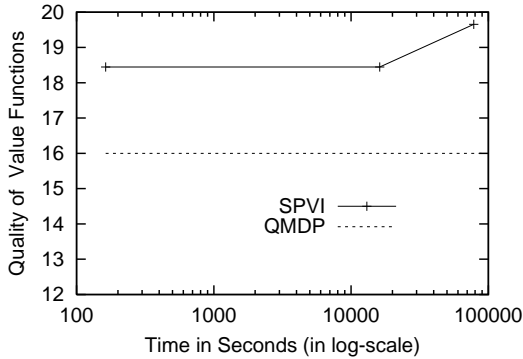

Figure 7: Performance of SPVI on office navigation problem

For reference, Table 4 gives detailed statistics on the number of the histories, iterations and vectors after each subset expansion. A note is about the number of iterations in the third column: when conducting value iteration over subsets, we also use the point-based improvements (Zhang, 2001). In the column, the number of point-based steps are excluded. The fourth column about the number of vectors provides some idea of why SPVI takes long time for this problem. This is because it generates a great number of vectors. After the third expansion, it uses 7,225 vectors to represent a value function over the belief subset.

| $i$-step expansion | histories# | iterations# | vectors# | rewards | time |
|---|---|---|---|---|---|
| 0 | 53 | 4 | 464 | 18.44 | 163 |
| 1 | 86 | 6 | 5178 | 18.44 | 16157 |
| 2 | 131 | 7 | 7225 | 19.65 | 78183 |

Table 4: Statistics on HKUST-CSD environment for SPVI

## 7. Related Work

In this paper, we propose restricted value iteration algorithms to accelerate value iteration for POMDPs. Two basic ideas behind restricted value iterations are (1) reducing the complexity of DP updates and (2) reducing the complexity of value functions. In this section, we discuss related work under these two categories. In addition, we give an overview on special POMDPs in the literature and the algorithms exploiting their problem characteristics.

## 7.1 Reducing the Complexity of DP Updates

In a broad sense, approaches reducing the complexity of DP updates can be roughly categorized into two classes: approaches conducting value updates over a (stationary) belief subset and approaches conducting value updates over a growing subset, although the boundary between these two classes is ambiguous in some cases.

The first class includes a family of grid-based algorithms, algorithms based on reachability analysis, algorithms using state-space decomposition and others. Grid-based algorithms update values for a finite grid and extrapolate values for non-grid belief states (Lovejoy, 1991; Hauskrecht, 1997; Zhou & Hansen, 2001). However, to guarantee optimality, the grid size is often exponential in the dimension of the state space. To tackle POMDPs with large state spaces, reachability analysis is a generally applicable technique. If an agent is informed of its initial belief, all belief states it can encounter form a finite set in case of a finite decision horizon. These belief states can be structured in a decision tree or AND/OR tree (Washington, 1997; Hansen & Ziberstein, 1998; Hansen, 1998; Bonet & Geffner, 2000). Although sometimes near optimality can be achieved at the initial belief state (Hansen, 1998), the algorithms in the cited articles cannot be applied to the case with unknown initial belief. State-space decomposition is an effective way to alleviate the curse of dimensionality. This approach has been successfully applied to MDPs (Dean & Lin, 1995; Dean, Givan, & Kim, 1998; Parr, 1998; Koller & Parr, 2000). Typically, to solve an MDP, one solves a number of small MDPs and uses their solutions to approximate that of the original MDP. However, the state-space decomposition approach cannot directly generalize to the POMDP context because of the inherent difficulty incurred by the continuum of the belief space.

Our theory and algorithms on restricted value iteration have significant differences from the above approaches. Through a well chosen belief subset, restricted value iterations can achieve convergence and optimality. It differs from grid-based algorithms in that it computes vector-based representations of value functions. Despite this difference, it is possible that grid-based algorithms can benefit from our theory on belief subset selection. This possibility is yet to be investigated. For instance, while choosing grid points, one should choose those within the belief subset $\tau(\mathcal{B})$. The reason follows. Since the belief states outside the set are never reachable, their values do not directly contribute to value updates for beliefs in the grid. With regard to the differences between the aforementioned algorithms and ours, our approach has no assumption about an agent's initial belief, although the belief subset is chosen via reachability analysis. Our algorithm differs from the decomposition techniques in that it solves a reformulated MDP instead of a set of small MDPs.

Approaches conducting value updates for a growing belief subset include real-time dynamic programming (RTDP) in the POMDP context (Barto, Bradtke, & Singh, 1995; Geffner & Bonet, 1998), a synthetic projection algorithm (Drummond & Bresina, 1990) and the envelope algorithm for Plexus planner in the MDP context (Dean et al., 1993). Naturally, they run as anytime algorithms. In RTDP, value updates are carried out for a belief subset, which grows as an agent explores the belief space. The main difference between SPVI and the above algorithms is how they expand the belief/state subset and how they choose beliefs/states for value updates. For subset expansion, SPVI adds more belief simplices, which often contains an infinite number of belief states, while the above algorithms mostly add a finite number of belief states. (It is also noted that reachability analysis is used in these expansions.) For value updates, SPVI improves values for the entire belief subset, while the above algorithms typically select a limited number of beliefs or states in the current subset.

## 7.2 Reducing the Complexity of Value Functions

Another idea behind restricted value iteration is concerned with the representational complexity of value functions. Intuitively, the representing set of a value function over a belief subset contains fewer vectors than that of the same value function over the belief space. This fact has been observed (Boutilier & Poole, 1996; Hauskrecht & Fraser, 1998), where POMDPs are represented compactly. When states are depicted by a set of variables, they are classified into observable variables and hidden variables. It is also noted that some belief states cannot be reached for certain combinations of observable variables and hidden variables. This fact has been exploited in approximating the solution to a medical treatment example (Hauskrecht & Fraser, 1998). Recent work along this thread includes a state-space compression technique exploiting the representational advantage (Poupart & Boutilier, 2002), and a technique of Principle Component Analysis (PCA) aiming at reducing the complexity of value functions (Roy & Gordon, 2002). However, it is unclear whether it is feasible to combine state-space compression with subset value iteration before we know how to conduct value iteration over a belief space induced by a compressed state space.

## 7.3 Solving Special POMDPs

Since solving a POMDP generally is computationally intractable, it is advisable to study POMDPs with special characteristics. The hope is that these characteristics may be exploited to find their near optimal solutions more efficiently. Special POMDPs examined in the literature include regional-observable POMDPs (Zhang & Liu, 1997), memory-resetting and discernible POMDPs (Hansen, 1998), even-odd POMDPs (Zubek & Dietterich, 2000) and generalized near-discernible POMDPs (Zhang, 2001). Interestingly, these POMDPs assume the existence of informative actions or observations such that somehow an agent is able to get more information about the world. In the following, we briefly discuss the assumptions behind informative POMDPs and near-discernible POMDPs and review existing work closely related to them. Before concluding this subsection, we also mentioned a couple of extensions to our current work.

An informative POMDP assumes that any observation restricts the world into a small set of states. This assumption is validated by a few problem instances with compact representations of state space. In the literature, some POMDP examples are actually informative POMDPs. One example is the *slotted Aloha* protocol problem (Bertsekas & Gallagher, 1995; Cassandra, 1998a), where the state of the system consists of the number of backlogged messages and the channel status. The channel status is observable and its possible assignments form the observation space. However, the system has no access to the number of backlogged messages. If the maximum number of backlogged messages is set to $m$ and there are $n$ possible values for the channel status, the number of states is $m \cdot n$. A particular assignment of channel status will restrict the system into $m$ states out of $m \cdot n$. A similar problem characteristic also exists in a non-stationary environment model proposed for reinforcement learning (Choi, Yeung, & Zhang, 1999).

A regional observable POMDP assumes that at any point in time an agent is restricted to a handful of world states (Zhang & Liu, 1997). The assumption leads to a value iteration algorithm that works with a belief subset and also exploits the low dimensional representation of vectors. We used the algorithm to solve informative POMDPs. However, we would like to discuss several differences. First, conceptually the assumptions are different for the two POMDP classes. In regional observable POMDPs, when the agent is restricted to a set of states (i.e., a *region*), the states in such a set are geometrically neighboring ones. However, in informative POMDPs, when the agent is restricted to a set of states, the states in a set are obtained by formally analyzing the observation model of the POMDP. It is possible that the states in the set are spatially distant from one another. Second, the algorithms for the two POMDP classes work in a quite different way. To ease presentation, we use `infoVI` and `roVI` to respectively denote our value iteration for informative POMDPs and that for regional observation POMDPs. In `infoVI`, the number of state sets is the product of the number of actions and the number of observations, while in `roVI`, the number of regions is subjectively chosen. In addition, the observations in `roVI` are *augmented*. An *augmented observation* consists of an original observation and a specific region. So, the number of augmented observations is the product of the number of original observations and the number of regions. Hence, `roVI` has to account for many more observations than `infoVI`. This fact is useful when comparing the efficiency of `infoVI` and `roVI`. Imagine what happens if `roVI` works with the region system, which consists of the state sets defined by `infoVI`, for an informative POMDP. Because `infoVI` accounts for fewer observations than `roVI`, it should be more efficient. Finally, the quality of the value function returned by `infoVI` is guaranteed for the entire belief space when it terminates with the strict stopping criterion. However, the quality of the value function by `roVI` in its original description is problematic even for the considered belief subset.

The other POMDP class examined in this paper is near-discernible POMDPs. A near discernible POMDP assumes that the actions are classified into information-rich ones and information-poor ones. The assumption is reasonable in several realistic domains. The first domain is the path planning problems (Cassandra, 1998a). The actions are categorized into goal-achieving and information-gathering ones, as discussed earlier. Another application domain is machine maintenance problems (Smallwood & Sondik, 1973; Hansen, 1998), where an agent usually can execute the following set of actions: manufacture, examine, inspect and replace. Among these actions, "inspect" is information-rich and the remaining three actions are information-poor.

A near discernible POMDP is a generalization of a memory-resetting (discernible) POMDP, which assumes that there exists actions resetting the world to an unique state. If such actions are performed, the agent knows that the world must be in a definite state. If the initial belief state is known and an optimal policy must execute one of such actions periodically, the number of belief states that the agent visits is finite. Accordingly, DP updates over a finite set of beliefs are much cheaper. However, after the discernibility assumption is relaxed, the agent may visit an infinite number of states and DP updates become more expensive. We therefore developed an anytime algorithm seeking a tradeoff between the solution quality and the size of the belief subset.

We also experimented with one extension of using SPVI to approximate the solutions of more general POMDPs (Zhang, 2001). The approximation scheme employs a thresholding technique. Given a POMDP and a threshold, a POMDP can be transformed to a new one, which differs from the original one in the observation model. The observation model in the transformed POMDP is obtained by ignoring the probabilities (in the original model) less than the threshold [7]. If the transformed POMDP is near discernible, its solution can be found by SPVI and be used to approximate that of the original POMDP. We have designed another maze problem that has no informative action/observation pair and therefore is expected to be not amenable to SPVI (Zhang, 2001). However, the transformed POMDP is amenable to SPVI. The experiments show that SPVI can quickly find a high quality solution for the transformed POMDP. In another case, if the transformed POMDP is informative, the algorithm exploiting low dimensional representations for informative POMDPs can be applied.

## 8. Conclusions

In this paper, we studied value iterations working with belief subset. We applied reachability analysis to select a particular subset. The subset is (1)*closed* in that no actions can lead the agent to belief states outside it; (2)*sufficient* in that value function defined over it can be extended into the belief space; and (3)*minimal* in that value iteration needs to consider at least the subset if it intends to achieve the quality of value functions. That the subset is closed enables one to formulate a subset MDP. We addressed the issues of representing the subset and pruning a set of vectors w.r.t. the subset. We then described the subset value iteration algorithm. For a given POMDP, whether the subset is proper can be determined *a priori*. If this is the case, subset value iteration carries the advantages of representation in space and efficiency in time. We also studied informative POMDPs and near-discernible POMDPs. For informative POMDPs, there are natural belief subsets for value iteration to work with. For near-discernible POMDPs, we developed an anytime value iteration algorithm seeking a tradeoff between the policy quality and the size of belief subsets.

## Acknowledgments

Research was partially supported by Hong Kong Research Grants Council under grant HKUST6088 / 01E. The authors thank Tony Cassandra and Eric Hansen for sharing with us their programs. The first author would like to thank Eric Hansen for in-depth discussions on belief subset selection and low dimensional representation, and Judy Goldsmith for valuable comments on an earlier writeup of the ideas in this paper. We are also grateful for the three anonymous reviewers who provided insightful comments and suggestions on this paper.

## Appendix A. Proofs

**Theorem 2** *For any pair $[a, z]$, the subset $\tau(\mathcal{B}, a, z)$ is a simplex.*

---

7. To complete the definition of the approximate observation model, one needs to re-normalize model parameters such that for an action and a state, the probabilities for all observations sum up to 1.0.

**Proof:** Suppose $b_i$ is a belief state such that $b_i(s) = 1.0$ for $s = s_i$ and 0 otherwise. It can be seen that $\{b_1, b_2, \cdots b_n\}$ is a basis of belief space $\mathcal{B}$. Each belief state $b(= (b(s_1), b(s_2), \cdots, b(s_n))$ can be represented as $\sum_{i=1}^{n} b(s_i) b_i$.

Let $k$ be the cardinality of the set $\{\tau(b_i, a, z) | P(z | b_i, a) > 0\}$. Without loss of generality, we enumerate the set as $\{\tau(b_1, a, z), \cdots, \tau(b_k, a, z)\}$. It suffices to show that $\tau(\mathcal{B}, a, z) = \Psi(\tau(b_1, a, z), \tau(b_2, a, z), \cdots, \tau(b_k, a, z))$. To prove it, we prove:

(1) $\tau(\mathcal{B}, a, z) \subseteq \Psi(\tau(b_1, a, z), \tau(b_2, a, z), \cdots, \tau(b_k, a, z))$ and

(2) $\Psi(\tau(b_1, a, z), \tau(b_2, a, z), \cdots, \tau(b_k, a, z)) \subseteq \tau(\mathcal{B}, a, z)$.

First, we prove (1). It suffices to show that any belief state $b'$ in $\tau(\mathcal{B}, a, z)$ must belong to the simplex $\Psi$. Since $b'$ is in $\tau(\mathcal{B}, a, z)$, there must exist a belief state $b$ in $\mathcal{B}$ such that $b' = \tau(b, a, z)$. We define a few constants as follows.

- For any $i \in \{1, \cdots, k\}$, $C_{b_i}$ is the probability of observing $z$ when action $a$ is executed in belief state $b_i$. Formally, $C_{b_i} = \sum_{s', s} P(z | s', a) P(s' | s, a) b_i(s)$.

- $C_b$ is the probability of observing $z$ when action $a$ is performed in $b$. Formally, $C_b = \sum_{s', s} P(z | s', a) P(s' | s, a) b(s)$.

- For any $i \in \{1, \cdots, k\}$, define $\lambda_i = C_{b_i} / C_b$.

Given these constants, we are going to prove $b' = \sum_i \lambda_i \tau(b_i, a, z)$. If this is true, i.e., $b'$ can be represented as a convex combination of the vectors in the basis, (1) is proven.

We start from $b' = \tau(b, a, z)$. If $\tau(b, a, z)$ is replaced by its definition, for a state $s'$,

$$b'(s') = \frac{1}{C_b} \sum_s P(z | s', a) P(s' | s, a) b(s)$$

By the definition of belief state $b_i$, we can rewrite the above equation as

$$b'(s') = \frac{1}{C_b} \sum_s \sum_{i \in \{1, \cdots, k\}} P(z | s', a) P(s' | s, a) b_i(s).$$

Trivially,

$$b'(s') = \frac{1}{C_b} \sum_s C_{b_i} \frac{\sum_i P(z | s', a) P(s' | s, a) b_i(s)}{C_{b_i}}.$$

By the definition of $\tau(b_i, a, z)$, rewriting the above equation, we have

$$b'(s') = \sum_i (\frac{C_{b_i}}{C_b}) \tau(b_i, a, z)(s').$$

By the definition of $\lambda_i$, the above equation yields

$$b'(s') = \sum_i \lambda_i \tau(b_i, a, z)(s').$$

If $b'$ and $\tau(b_i, a, z)$ are regarded as column vectors, the above equation means

$$b' = \sum_i \lambda_i \tau(b_i, a, z).$$

Therefore, we prove that if there is a belief state $b$ such that $b' = \tau(b, a, z)$, $b'$ can be represented as a convex combination of the vectors in the basis. This means $b'$ must be in the simplex $\Psi$.

To prove (2), we prove that any belief state $b'$ in the simplex $\Psi$ must be in the subset $\tau(\mathcal{B}, a, z)$. It suffices to show that there exists a belief state $b$ in $\mathcal{B}$ such that $b' = \tau(b, a, z)$.

Since $b'$ is in $\Psi$, there must exist a set of nonnegative $\lambda_i$s such that $b' = \sum_{i=1}^{k} \lambda_i \tau(b_i, a, z)$. If we replace $\tau(b_i, a, z)$ by its definition, then: for a state $s'$,

$$b'(s') = \sum_i \lambda_i \frac{\sum_s P(z|s', a) P(s'|s, a) b_i(s)}{\sum_{s', s} P(z|s', a) P(s'|s, a) b_i(s)}.$$

If we denote $\sum_{s', s} P(z|s', a) P(s'|s, a) b_i(s)$ by a constant $C_{b_i}$, then

$$b'(s') = \sum_i \sum_s P(z|s', a) P(s'|s, a) b_i(s) \frac{\lambda_i}{C_{b_i}}.$$

Exchanging the summation order of $i$ and $s$ and making use of the definition of $b_s(i)$, we have

$$b'(s') = \sum_s \frac{\lambda_s}{C_{b_s}} P(z|s', a) P(s'|s, a).$$

We define a belief state $b$ as follows: for any $s$,

$$b(s) = \frac{\lambda_s / C_{b_s}}{\sum_s \lambda_s / C_{b_s}}.$$

It can be seen that

$$b'(s') = \frac{\sum_s P(z|s', a) P(s'|s, a) b(s)}{\sum_{s', s} P(z|s', a) P(s'|s, a) b(s)}.$$

Therefore, we proved that for any $b'$ in $\Psi$ there exists a belief state $b$ such that $b' = \tau(b, a, z)$. Consequently, $b' \in \tau(\mathcal{B}, a, z)$. □

**Theorem 6** *If $\max_{b \in \tau(\mathcal{B})} |V_n^{\tau(\mathcal{B})}(b) - V_{n-1}^{\tau(\mathcal{B})}(b)| \leq \epsilon(1 - \lambda)/(2\lambda^2 |\mathcal{Z}|)$, then the $V_{n-1}^{\tau(\mathcal{B})}$-improving policy is $\epsilon$-optimal over the entire belief space $\mathcal{B}$.*

**Proof:** It suffices to show $\max_{b \in \mathcal{B}} |V_{n+1}(b) - V_n(b)| \leq \epsilon(1 - \lambda)/(2\lambda)$. For any $b \in \mathcal{B}$,

$$
\begin{aligned}
& |V_{n+1}(b) - V_n(b)| \\
= \ & |\max_a \{r(b, a) + \lambda \sum_z V_n^{\tau(\mathcal{B}, a, z)}(\tau(b, a, z))\} - \max_a \{r(b, a) + \lambda \sum_z V_{n-1}^{\tau(\mathcal{B}, a, z)}(\tau(b, a, z))\}| \quad (1) \\
\leq \ & |r(b, a^*) + \lambda \sum_z V_n^{\tau(\mathcal{B}, a^*, z)}(\tau(b, a^*, z)) - r(b, a^*) - \lambda \sum_z V_{n-1}^{\tau(\mathcal{B}, a^*, z)}(\tau(b, a^*, z))| \quad (2) \\
\leq \ & |\lambda \sum_z (V_n^{\tau(\mathcal{B}, a^*, z)}(\tau(b, a^*, z)) - V_{n-1}^{\tau(\mathcal{B}, a^*, z)}(\tau(b, a^*, z)))| \quad (3) \\
\leq \ & \lambda |\mathcal{Z}| \max_z |V_n^{\tau(\mathcal{B}, a^*, z)}(\tau(b, a^*, z)) - V_{n-1}^{\tau(\mathcal{B}, a^*, z)}(\tau(b, a^*, z))| \quad (4) \\
\leq \ & \lambda |\mathcal{Z}| \epsilon(1 - \lambda)/(2\lambda^2 |\mathcal{Z}|) \quad (5) \\
\leq \ & \epsilon(1 - \lambda)/(2\lambda) \quad (6)
\end{aligned}
$$

where

- At Step (1), value functions are replaced by their definitions;

- At Step (2), $a^*$ is the $V_n^{\tau(\mathcal{B})}$-improving action for $b$ but it is not necessarily $V_{n-1}^{\tau(\mathcal{B})}$-improving;

- At Step (5), the given condition is used;

- Other steps are trivial. $\qquad\square$

## Appendix B. Informative POMDPs: An Elevator Problem

This appendix describes a 96-state informative POMDP and empirical results of value iteration over $\phi(\mathcal{B})$. The problem is adapted from existing research (Choi et al., 1999). Our purpose is to show that restricted value iteration is able to solve larger POMDPs than standard value iteration.

### Problem Formulation

An elevator operates in a two-floor residential building. There are three patterns on the passengers' arrival: high arrival rate in the first floor and low in the second floor; low arrival rate in the first floor and high in the second floor; equal arrival rates. As time varies from the morning to the night in a day, these patterns change according to a probability distribution. To keep track of the pick-up and drop-off requests, the elevator sets up four buttons in its control panel: two buttons record the pick-up and drop-off requests for the first floor, two other buttons keep the same information for the second floor. The elevator is also aware of which floor it is on. In order to fulfill the requests at a floor, the elevator first moves upwards or downwards so that it reaches the floor; then, the elevator stays at the floor until the passengers finish entering or exiting. The objective of the elevator is to minimize certain penalty or cost in the long run.

The problem can be formulated into a POMDP framework. A state consists of six components: the arrival pattern, the pick-up requests for two floors, the drop-off requests for two floors and the elevator's position. We use six variables to denote the components respectively. A state is an assignment to all the variables. The arrival pattern takes on three possible values for three different patterns. If there are passengers waiting in the lobby of the first floor, the pick-up request is set; otherwise, it is unset. If there are passengers in the elevator intending to get off at the first floor, the drop-off request for the first floor is set; otherwise, it is unset. Similarly, for the second floor, the variables for the pick-up/drop-off requests can be set accordingly. If the elevator is at the first floor, its position is set to first; for the second floor, its position is set to second. The number of states is $3*2*2*2*2 = 96$. Each observation has five components; it has the same components as a state except the arrival pattern. There are as many as $2*2*2*2*2 = 32$ observations. The elevator may execute one of three actions, namely `go.up`, `go.down` and `stay`. The restriction is, when it is at the first floor, it cannot perform `go.down`; when it is at the second floor, the action `go.up` cannot be performed.

The uncertainty stems from the probabilities of the changes of the arrival patterns. When the elevator executes `go.up`, each component evolves as follows. The arrival pattern changes according to a predetermined probability distribution. The components for pickup/drop-off requests remain. The position changes from first to second. The effects of the action `go.down` can be described similarly. When the elevator performs the action `stay`, the arrival pattern changes similarly. All the requests at the floor are fulfilled and the corresponding variables are reset. For instance, if a passenger would like to get off at the first floor, when the elevator at the first floor performs stay, the passenger is able to get off. We say that the elevator fulfills the drop-off requests for the first floor. For another instance, if passengers like to enter the elevator at the second floor, they can do so only when the elevator performs action `stay` at the second floor. We say the pickup request at the second floor is fulfilled in this case. It is also allowable for the elevator to fulfill two requests at one time point. For example, if there are both pick-up and drop-off requests at the first floor, when the elevator performs action `stay`, the passengers can enter and exit within one time point. We say it fulfills two requests. Note that only when an action `stay` is performed, the elevator can fulfill its request. Since the variable of arrival pattern changes at any time moment, the elevator changes its states probabilistically after performing any action.

The elevator is informed of partial knowledge of its state transition. After the elevator performs an action, it knows the changes of components of its states: variables pick-up and drop-off for each floor and its position. However, since it does not know the arrival pattern and it is a component of the state, the observations cannot reveal the identities of the states. This is the partial observability. However, since there are only three possible arrival patterns, each observation reveals that the elevator must be in only three possible states. Therefore, the POMDP is informative.

The performance of the elevator can be measured in different ways for diverse applications. We define a measure to minimize the unsatisfactory degree of the service the elevator provides. We encode this in our reward model. At any time point, the elevator serves one of four requests: pick-up requests at the first/second floor, drop-off requests at the first/second floor. After performing an action, if any of these 4 requests is unfulfilled, the elevator receives a penalty of 0.25. For instance, if the elevator un-fulfills either the pick-up or drop-off request(if they are set) at the first floor, it receives a penalty of $0.25 * 2 = 0.5$. The objective of the elevator is to minimize total discounted penalty in a long run.

For convenience, we use A.$i$ to denote arrival patterns for $i = 1, 2, 3$. In our experiments, the transition probabilities are set as in the following table. Basically, each pattern remains fixed with probability 0.90 and changes to another with 0.05.

|      | A.1  | A.2  | A.3  |
|------|------|------|------|
| A.1  | 0.90 | 0.05 | 0.05 |
| A.2  | 0.05 | 0.90 | 0.05 |
| A.3  | 0.05 | 0.05 | 0.90 |

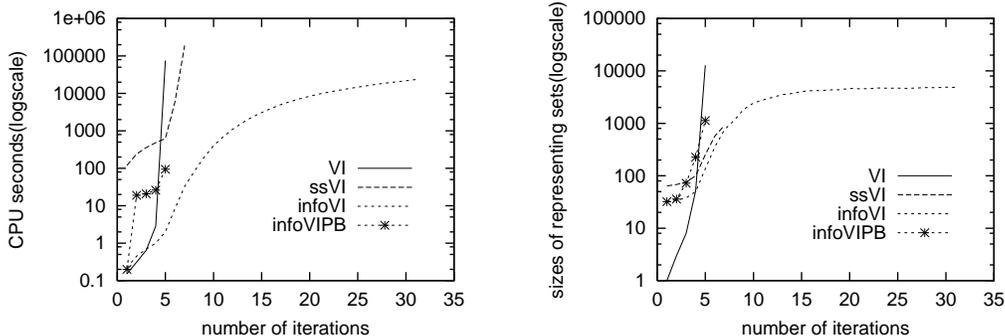

Figure 8: Performance of `VI`, `ssVI`, `infoVI` and `infoVIPB` on Elevator

## Empirical Studies

We collect the time costs and the actual number of vectors generated at each iteration of algorithms `VI`, `ssVI`, `infoVI` and `infoVIPB` referring to `infoVI` integrated with a point-based procedure (Zhang, 2001). The results are presented in Figure 8.

The first chart in the figure shows the time costs against the iterations. The algorithms are set to compute a 0.1-optimal value function. For `infoVIPB`, we exclude the iterations for point-based improvements. Overall, we see that `VI` and `ssVI` by no means can solve the problem, `infoVI` is likely to solve it given sufficient time and `infoVIPB` is able to solve it easily. When `infoVIPB` runs, it uses the loose stopping criterion. This is because if the strict one is used, the threshold is close to the round-off precision parameter.

For the first seven iterations, `ssVI` takes 190,000 seconds, `infoVI` only 32 seconds. The performance difference is drastic. As `infoVI` proceeds, it takes about 1,100 seconds for one iteration. It is evident that `infoVI` is able to compute a near optimal value function if given sufficient time. When the point-based technique is integrated, `infoVIPB` is able to terminate in 94 seconds after five steps of DP updates over $\phi(\mathcal{B})$. Since most of these algorithms cannot terminate within a reasonable time limit, we compare the data on the 6th iteration among them. This is the last iteration we are able to gather statistics for `VI`. For this iteration, `VI` takes 76,000 seconds, `ssVI` 6,000 seconds, `infoVIPB` only 8 seconds.

The second chart in Figure 8 depicts the number of vectors generated at each iteration for the tested algorithms. For `ssVI`, we collect the sum of the numbers of vectors representing value functions over $|\mathcal{A}| \cdot |\mathcal{Z}|$ $\tau$-simplices. For `infoVI` and `infoVIPB`, we collect the sum of the numbers of representing vectors for $|\mathcal{Z}|$ $\phi$-simplices. This is because for this problem the observation models are independent of the actions.

From the chart, we see that `VI` generates significantly more vectors than `ssVI` and `infoVI`. In our experiments, after `infoVIPB` terminates, it produces 1,132 vectors. For the same reason as above, we compare the numbers of the vectors after the 6th iterations for these algorithms. After the iteration, `VI` generates 12,000 vectors. For `ssVI` and `infoVI`, this number is 252 and 136 respectively. As DP updates proceed, it is conceivable that the number of vectors generated by `VI` will increase sharply and hence the DP updates are extremely inefficient. For `infoVIPB`, since the final number of generated vectors are rather small, together with the fact that point-based improvement effectively reduces the number of iterations over the $\phi(\mathcal{B})$, it is possible to compute a near optimal value function within a rather small time limit as it turns out.